\documentclass{article}

\PassOptionsToPackage{numbers}{natbib}
\usepackage[preprint]{neurips_2026}

\usepackage[utf8]{inputenc} % allow utf-8 input
\usepackage[T1]{fontenc}    % use 8-bit T1 fonts
\usepackage[dvipsnames]{xcolor}         % colors

\definecolor{olive}{rgb}{0.6, 0.6, 0.2}
\definecolor{sand}{rgb}{0.8666666666666667, 0.8, 0.4666666666666667}
\definecolor{wine}{rgb}{0.5333333333333333, 0.13333333333333333, 0.3333333333333333}
\definecolor{deblue}{RGB}{11,132,147}
\definecolor{ocra}{RGB}{204, 119, 34}

\usepackage{microtype}
\usepackage{graphicx}
\usepackage{subcaption}
\usepackage{booktabs} % for professional tables
\usepackage{url}            % simple URL typesetting
\usepackage{amsfonts}       % blackboard math symbols
\usepackage{nicefrac}       % compact symbols for 1/2, etc.

\usepackage{hyperref}

\usepackage{algorithm}
\usepackage{algorithmic}

\usepackage{amsmath}
\usepackage{amssymb}
\usepackage{mathtools}
\usepackage{amsthm}
\usepackage{enumitem}

\usepackage[capitalize,noabbrev]{cleveref}

\theoremstyle{plain}

\theoremstyle{definition}

\theoremstyle{remark}

\usepackage[disable,textsize=tiny]{todonotes}
\usepackage{multirow}
\usepackage{tikz}

\title{Posterior Inference in Latent Space \\
for Scalable Constrained Black-box Optimization}

\author{%
  Kiyoung Om$^{1*}$ \quad Kyuil Sim$^{1*}$ \quad Taeyoung Yun$^{1}$\thanks{Equal contribution authors.} \quad Hyeongyu Kang$^1$ \quad Jinkyoo Park$^1$ \\
  $^1$Korea Advanced Institute of Science and Technology (KAIST)\\
  \texttt{\{se99an, kyuil.sim, 99yty, khg2000v, jinkyoo.park\}@kaist.ac.kr}
}

\begin{document}

\maketitle

\begin{abstract}
Optimizing high-dimensional black-box functions under black-box constraints is a pervasive task in a wide range of scientific and engineering problems. 
These problems are typically harder than unconstrained problems due to hard-to-find feasible regions. 
% In this paper, we propose a new framework to solve high-dimensional, constrained black-box optimization problems via posterior inference in the latent space of generative models.
In this work, we reformulate constrained black-box optimization as posterior inference, and perform this inference in the latent space of generative models.
Our method iterates through two stages. First, we train flow-based models to capture the data distribution and surrogate models that predict both function values and constraint violations. Second, we cast the candidate selection problem as a posterior inference problem to effectively search for promising candidates that have high objective values while not violating the constraints. 
Concretely, we utilize outsourced diffusion models to amortize the sampling from the posterior distribution in the latent space of flow-based models, which can bypass the issue of mode collapse. 
We empirically demonstrate that our method achieves superior performance across synthetic and real-world tasks. Our code is available \href{https://github.com/umkiyoung/CiBO}{here}.
\end{abstract}

\section{Introduction}

% Optimizing high-dimensional black-box functions under black-box constraints is a fundamental task across numerous scientific and engineering problems, including machine learning  \citep{gardner2014bayesian}, drug discovery \citep{griffiths2020constrained}, control \citep{berkenkamp2016safe}, and industrial design \citep{maathuis2024high}. In most cases, these problems are much harder than unconstrained problems due to analytically undefined and hard-to-find feasible regions \citep{eriksson2021scalable}.
Optimizing high-dimensional black-box functions under black-box constraints is a fundamental task across numerous scientific and engineering problems, including machine learning~\citep{gardner2014bayesian}, drug discovery~\citep{griffiths2020constrained}, control~\citep{berkenkamp2016safe}, and industrial design~\citep{maathuis2024high}. These problems are substantially harder than their unconstrained counterparts. Feasible regions are analytically undefined and often occupy only a small fraction of the search space~\citep{eriksson2021scalable}, and high dimensionality compounds all of these difficulties.

% Bayesian optimization (BO) has been widely used to solve black-box optimization problems in a sample-efficient manner \citep{kushner1964new,garnett2023bayesian}. While most BO methods focus on unconstrained optimization problems, some works address problems with black-box constraints by developing new acquisition functions \citep{gardner2014bayesian,hernandez2015predictive} or relaxing the constraints \citep{picheny2016bayesian,JMLR:v20:18-227}.
% However, even without constraints, BO methods scale poorly to high dimensionality \citep{eriksson2019scalable}. Moreover, incorporating constraints makes the function landscape more complex, hindering accurate estimation of surrogate models.
Bayesian optimization (BO) has been widely used to solve black-box optimization problems in a sample-efficient manner~\citep{kushner1964new,garnett2023bayesian}. Some BO methods extend to black-box constraints by developing new acquisition functions~\citep{gardner2014bayesian,hernandez2015predictive} or relaxing them via Lagrangian methods~\citep{picheny2016bayesian,JMLR:v20:18-227}. However, BO inherently scales poorly to high dimensionality even without constraints~\citep{eriksson2019scalable}, and incorporating constraints further complicates the function landscape, hindering accurate estimation of surrogate models.

% Recently, generative models have emerged as a promising approach for constrained black-box optimization problems \citep{kong2025diffusion,xing2025black,uehara2025reward}. Instead of searching for candidates that maximize the acquisition function, we can capture the distribution of promising candidates with generative models to explore the search space more effectively. However, when it comes to high-dimensional problems with a large number of constraints, training generative models to match such a distribution is vulnerable to the mode collapse as the target distribution is highly complex with distinct modes \citep{venkatraman2024amortizing,uehara2024fine, domingo-enrich2025adjoint}.
Generative models have recently emerged as a promising alternative for optimization problems with black-box constraints~\citep{kong2025diffusion,xing2025black,uehara2025reward}. By capturing the distribution of promising candidates, they can explore the search space more flexibly than solving acquisition function maximization \citep{yun2025posterior}. However, in high-dimensional constrained spaces, the target distribution over feasible and high-scoring candidates is highly multi-modal and exhibits large flat regions, making sampling from this complex distribution intractable. One can approximate it via Markov Chain Monte Carlo (MCMC), but such methods scale poorly to high-dimensional spaces~\citep{chung2023diffusion, wu2023practical}. Fine-tuning approaches to match the target distribution are prone to mode collapse when the distribution is highly multi-modal~\citep{venkatraman2024amortizing, venkatraman2025outsourced}.

\begin{figure}[t]
    \centering
    \vspace{-5pt}
    \includegraphics[width=0.95\textwidth]{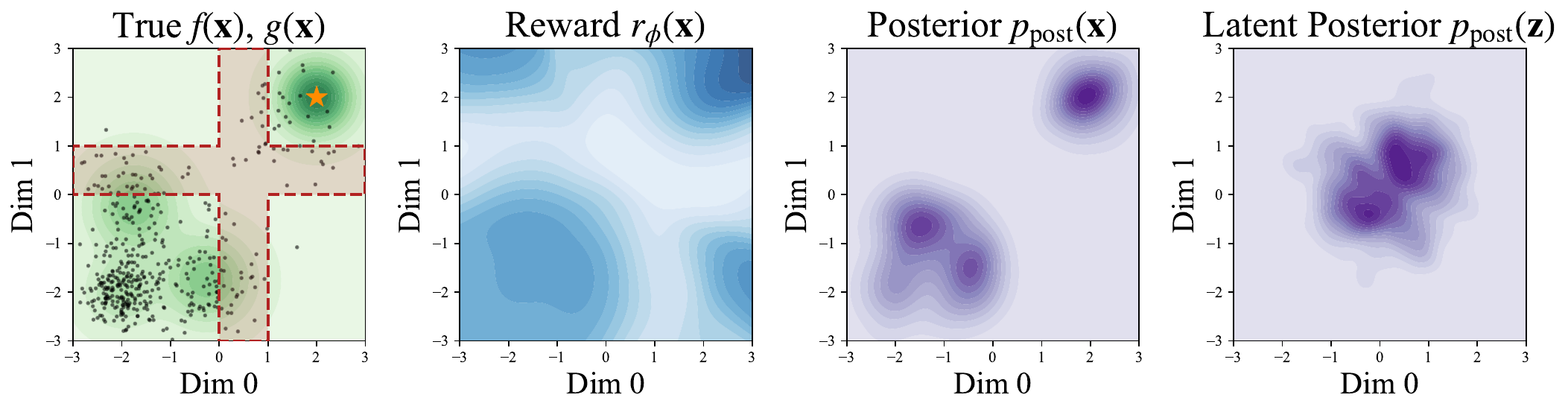}
    \vspace{-5pt}
    \caption{Motivating Figure (\textcolor{orange}{$\bigstar$}: global optimum, \textcolor{red}{\texttt{-\,-\,-}}: infeasible region). In constrained optimization problems, the posterior distribution in the data space, $p_{\mathrm{post}}(\mathbf{x})$, is highly multi-modal since the constraint penalty makes a complex reward landscape. In latent space, $p_{\mathrm{post}}(\mathbf{z})$ is smoother and more amenable to amortized inference, enabling robust exploration towards promising regions.}
    \label{fig:motivation}
    \vspace{-15pt}

\end{figure}

% In this paper, we propose a novel generative model-based framework for constrained black-box optimization to overcome the aforementioned limitations.
% We first define a distribution of promising candidates as a posterior distribution, which can be constructed by multiplying the data distribution by a Lagrangian-relaxed objective.
% To sample candidates from the posterior distribution, our key idea is to amortize inference in the latent space of a generative model that approximates the data distribution, as illustrated in \Cref{fig:motivation}. Since the posterior distribution in the latent space is much smoother than that in the data space, we can approximate the distribution more accurately and alleviate the mode collapse problem \citep{coeurdoux2023normalizing}.
In this paper, we propose \textbf{CiBO}, a novel framework for scalable constrained black-box optimization that overcomes the above limitations. We formalize candidate selection as sampling from the posterior distribution $p_{\text{post}}(\mathbf{x}) \propto p_\theta(\mathbf{x})\exp(\beta \cdot r_\phi(\mathbf{x}))$, where $p_\theta$ is a learned flow-based prior and $r_\phi$ is a proxy that encodes both high objective values and constraint feasibility. Our key observation is that while the posterior has distinct modes in data space, the corresponding posterior in the latent space is significantly smoother, as illustrated in \Cref{fig:motivation}. Therefore, we amortize posterior inference in the latent space and project latents into the data space rather than fine-tuning flow-based models.
% Amortizing inference in this latent space is far more tractable and alleviates the mode collapse problem~\citep{venkatraman2025outsourced,coeurdoux2023normalizing}.

% Our method iterates through two stages. First, we train a flow-based model to capture the current data distribution. We also train surrogate models to predict the objective value and constraints, respectively. For the surrogate models, we use an ensemble of neural networks to quantify the uncertainty of the prediction, as we have only a small amount of data that covers a tiny fraction of the whole search space.
% Second, we sample candidates from the posterior distribution, where we treat a trained flow-based model as a prior, and a Lagrangian-relaxed objective as a reward function.
% As the posterior distribution is highly multi-modal and has a large plateau due to constraints, especially when constraint feedback is given as binary indicators of feasibility, we train a diffusion sampler that amortizes inference of the posterior distribution in the latent space of flow models. Then, we sample latents from the diffusion sampler and project them into data space using a deterministic mapping derived from the trained flow model. By repeating these two stages, we can progressively move toward high-scoring regions while not violating the constraints.
Our method iterates through two stages. In \textbf{Phase 1}, we train a flow-based model on the current dataset via reweighting, alongside an ensemble of surrogate models with uncertainty quantification to predict objective values and constraints. In \textbf{Phase 2}, we sample candidates by training an outsourced diffusion sampler~\citep{venkatraman2025outsourced} that amortizes the posterior distribution in the latent space via the trajectory balance objective~\citep{malkin2022trajectory}. 
% Operating in the smooth latent space enables the sampler to capture multi-modal and flat target distributions without mode collapse. 
After sampling and filtering, we evaluate the selected candidates, update the dataset, and repeat until the evaluation budget is exhausted.

% We conduct extensive experiments on three synthetic and four real-world benchmarks to validate the superiority of our method on scalable constrained black-box optimization problems. We also consider a more challenging scenario where the feedback from the constraints is given as a binary value. We empirically show that our method outperforms several competitive baselines across different tasks.
We conduct extensive experiments on three synthetic and four real-world benchmarks, including a more challenging indicator-constraint setting where only binary feasibility feedback is available. We empirically show that CiBO outperforms several competitive baselines from BO to generative model-based approaches across different tasks. Beyond performance, our ablation studies reveal that amortized inference in the latent space enables robust exploration on feasible regions.

\section{Related Works}

\subsection{Constrained Black-box Optimization}
% Most scientific and engineering optimization problems involve black-box constraints, such as the synthesizability of molecules in chemical design \citep{griffiths2020constrained} and safety constraints in robot control policies \citep{berkenkamp2016safe}. Existing BO methods solve this problem by either integrating the constraints directly into the acquisition function (cEI~\citep{schonlau1998global}, LogcEI~\citep{ament2023unexpected}) or by employing trust region approaches for scalability (SCBO~\citep{eriksson2021scalable}, PCAGP-SCBO~\citep{maathuis2024high}). Another line of work utilizes evolutionary algorithms like CMA-ES~\citep{hansen2006cma, atamna2016augmented} with an augmented Lagrangian method to navigate constrained spaces. However, the performance of these methods often degrades as dimensionality and the number of evaluations increase, which motivates the need for a more scalable approach.
Most scientific and engineering optimization problems involve black-box constraints, such as the synthesizability of molecules in chemical design \citep{griffiths2020constrained} and safety constraints in robot control policies \citep{berkenkamp2016safe}.
% Various strategies have been proposed to handle such constraints in the BO literature.
To handle such constraints, various strategies have been proposed in the BO literature.
cEI \citep{schonlau1998global} and LogcEI \citep{ament2023unexpected} incorporate constraints violations into the acquisition function.
% In terms of the high-dimensional search spaces, SCBO \citep{eriksson2021scalable} proposes a trust-region approach with bilog transformations to address scalable constrained black-box optimization problems. PCAGP-SCBO \citep{maathuis2024high} extends this idea by applying  PCA \citep{jolliffe2002principal} to handle a large number of constraints.
To scale to high-dimensional search spaces, SCBO \citep{eriksson2021scalable} proposes a trust-region approach with bilog transformations to address scalable constrained black-box optimization problems. PCAGP-SCBO \citep{maathuis2024high} extends this idea by applying PCA \citep{jolliffe2002principal} to handle a large number of constraints. FuRBO~\citep{ascia2025feasibility} proposes a feasibility-driven trust region approach that adapts the search region based on constraint satisfaction.

Another line of work integrates the objective function with constraints via augmented Lagrangian or the Alternating Direction Method of Multipliers (ADMM) \citep{picheny2016bayesian, JMLR:v20:18-227}, and solves unconstrained optimization problems by vanilla BO methods.
While our method also applies Lagrangian relaxation, we take a fundamentally different approach from standard BO literature in the candidate selection procedure. Instead of searching for inputs that maximize the acquisition function, we frame candidate selection as a posterior inference problem.

% Evolutionary algorithms can also be applied. CMA-ES~\citep{hansen2006cma,atamna2016augmented} employs continuous parameter adaptation and handles constraints either by assigning the objective values of infeasible points to zero or by applying the augmented Lagrangian.
Constrained black-box optimization has also been studied in evolutionary algorithms and operations research.
% Evolutionary algorithms also address constrained optimization. 
CMA-ES~\citep{hansen2006cma,atamna2016augmented} handles constraints either by assigning the objective values of infeasible points to zero, while COBYLA\citep{powell1994direct} maintains trust regions to perform local search in feasible regions.
% Despite these advancements, we observe severe performance degradation as dimensionality increases, indicating that a new scalable algorithm should be proposed.

\subsection{Generative Model-based Optimization}
% Motivated by the success of generative models \citep{ramesh2022hierarchical, esser2024scaling}, there are several attempts to utilize generative models to optimize black-box functions with~\citep{kong2025diffusion} and without constraints~\citep{krishnamoorthy2023diffusion, wu2024diffbbo, yun2025posterior}.
Motivated by the success of generative models \citep{ramesh2022hierarchical, esser2024scaling}, several methods have been proposed to leverage generative models for black-box optimization, both with and without constraints.

In offline black-box optimization, DDOM \citep{krishnamoorthy2023diffusion} trains a conditional diffusion model with classifier-free guidance~\citep{ho2022classifier} and applies loss reweighting to emphasize samples with high objective values. DiffOPT~\citep{
kong2025diffusion} formulates offline optimization as a constrained optimization problem. It applies diffusion to capture data distribution, followed by Langevin dynamics or an iterative importance sampling procedure to explore promising regions while not deviating too far from the data distribution.

% In online black-box optimization, DiffBBO \citep{wu2024diffbbo} and DiBO \citep{yun2025posterior} both leverage diffusion models and incorporate uncertainty estimation during candidate selection.
% DiBO treats candidate selection as posterior inference to guide sampling toward regions of high reward and uncertainty, while DiffBBO selects conditioning targets by employing an uncertainty-based acquisition function. Unfortunately, constrained black-box optimization in the online setting remains unexplored, where we need to actively explore a vast search space to discover feasible regions iteratively \citep{jain2022biological, kim2024improved}. In this paper, we propose an effective generative model-based method to solve high-dimensional constrained black-box optimization.
% DiBO also treats candidate selection as posterior inference, but operates in data space and is designed for unconstrained optimization. As a result, its fine-tuning-based approach is susceptible to mode collapse when the posterior becomes highly multi-modal and flat, which is the typical structure in constrained settings. DiffBBO selects conditioning targets by employing an uncertainty-based acquisition function and likewise does not address constraints. In this paper, we propose an effective generative model-based method to solve high-dimensional constrained black-box optimization.
In online black-box optimization, DiffBBO \citep{wu2024diffbbo} and DiBO \citep{yun2025posterior} both leverage diffusion models for candidate selection, but are designed to solve unconstrained optimization problems. Unfortunately, utilizing generative models for constrained black-box optimization in the online setting remains unexplored. In this paper, we propose an effective generative model-based method that performs posterior inference in the latent space to effectively search for candidates in feasible regions.
% There are several attempts to utilize generative models for black-box optimization.
% In an offline setting, DDOM \citep{krishnamoorthy2023diffusion} trains a conditional diffusion model with classifier-free guidance and applies a loss-reweighting to emphasize samples with high objective values. DiffOPT \citep{kong2025diffusion} solves a constrained optimization problem. It applies diffusion to capture data distribution, followed by an iterative importance‐sampling procedure. In an online setting, DiffBBO \citep{wu2024diffbbo} and citep \cite{yun2025posterior} both leverage diffusion models and incorporate uncertainty estimation during candidate selection. DiBO treats candidate selection as posterior inference to guide sampling toward regions of high reward and uncertainty, while DiffBBO selects conditioning targets by employing an uncertainty-based acquisition function.
% However, constrained black-box optimization in the online setting remains unexplored.

\subsection{Posterior Inference in Flow-based and Diffusion Models}
Given a flow-based or diffusion prior $p_\theta(\mathbf{x})$ trained on a dataset and a reward function $r(\mathbf{x})$, sampling from the unnormalized posterior $p_{\text{post}}(\mathbf{x}) \propto p_\theta(\mathbf{x}) r(\mathbf{x})$ has numerous applications in downstream tasks such as conditional image generation \citep{ho2022classifier,dhariwal2021diffusion}, Bayesian inverse problems \citep{chung2023diffusion,venkatraman2024amortizing}, and aligning pretrained models with human preference \citep{ fan2023dpok, venkatraman2025outsourced}.
However, direct sampling from this unnormalized posterior distribution $p_\text{post}(\mathbf{x})$ is generally intractable~\citep{venkatraman2024amortizing, feng2025on}.

% Given a generative model prior $p_\theta(\mathbf{x})$ trained on a dataset and a reward function $r(\mathbf{x})$, sampling from the posterior distribution $p_{\text{post}}(\mathbf{x}) \propto p_\theta(\mathbf{x}) r(\mathbf{x})$ has numerous applications in downstream tasks, including conditional image generation \citep{dhariwal2021diffusion, ho2022classifier}, Bayesian inverse problems \citep{venkatraman2024amortizing, song2022solving, chung2023diffusion}, and aligning pretrained models through human preference data \citep{domingo-enrich2025adjoint, venkatraman2025outsourced,fan2023dpok}. However, flow-based and diffusion models exhibit hierarchical structures within their generation processes, which make direct sampling from the unnormalized posterior $p_\theta(\mathbf{x})r(\mathbf{x})$ inherently intractable~\citep{feng2025on}.

To address this problem, some approaches train classifiers directly within intermediate noised spaces~\citep{dhariwal2021diffusion, lu2023contrastive} while others approximate posterior sampling via Markov Chain Monte Carlo (MCMC) procedures \citep{coeurdoux2023normalizing,chung2023diffusion, wu2023practical, cardoso2024monte}. However, these methods struggle to scale to high-dimensional settings.
Conversely, several methods use reinforcement learning~\citep{fan2023dpok, black2024training} or stochastic optimal control \citep{domingo-enrich2025adjoint} to amortize posterior sampling. These methods primarily fine-tune the parameters of the prior using on-policy samples. Unfortunately, naive implementations of fine-tuning methods can be prone to mode collapse when the target distribution is highly multi-modal and has a large flat region \cite{venkatraman2024amortizing, venkatraman2025outsourced}.

% On the other hand, several methods utilize reinforcement learning \citep{fan2023dpok, black2024training} or generative flow networks \citep{bengio2023gflownet} to fine-tune the diffusion models~\citep{venkatraman2024amortizing}. Furthermore, adjoint matching \citep{domingo-enrich2025adjoint} proposes fine-tuning of flow-based and diffusion models via a stochastic optimal control formulation to amortize the posterior distributions. Meanwhile, naive implementations of fine-tuning methods can be prone to mode collapse when the target distribution is highly multi-modal and has a large plateau region \citep{venkatraman2025outsourced}.

% Some approaches train classifiers directly within intermediate noised spaces \cite{dhariwal2021diffusion, lu2023contrastive} while others approximate posterior sampling via Markov Chain Monte Carlo (MCMC) procedures \cite{chung2023diffusion, song2023loss, cardoso2024monte, coeurdoux2023normalizing}.
% Additionally, reinforcement learning \cite{black2024training, fan2023dpok}, and generative flow networks \cite{bengio2023gflownet, venkatraman2024amortizing} have been utilized to fine-tune pretrained diffusion models. Adjoint-matching methods with Stochastic Optimal Control (SOC) formulations have been utilized to fine-tune pretrained flow-based models \cite{domingo-enrich2025adjoint} to amortize posterior distributions.

% Nevertheless, each of these methods presents limitations. Training classifiers in noisy data spaces and employing MCMC methods scales poorly to high-dimensionality.

\begin{figure}[t]
    \centering
    \vspace{-10pt}
    \includegraphics[width=\textwidth]{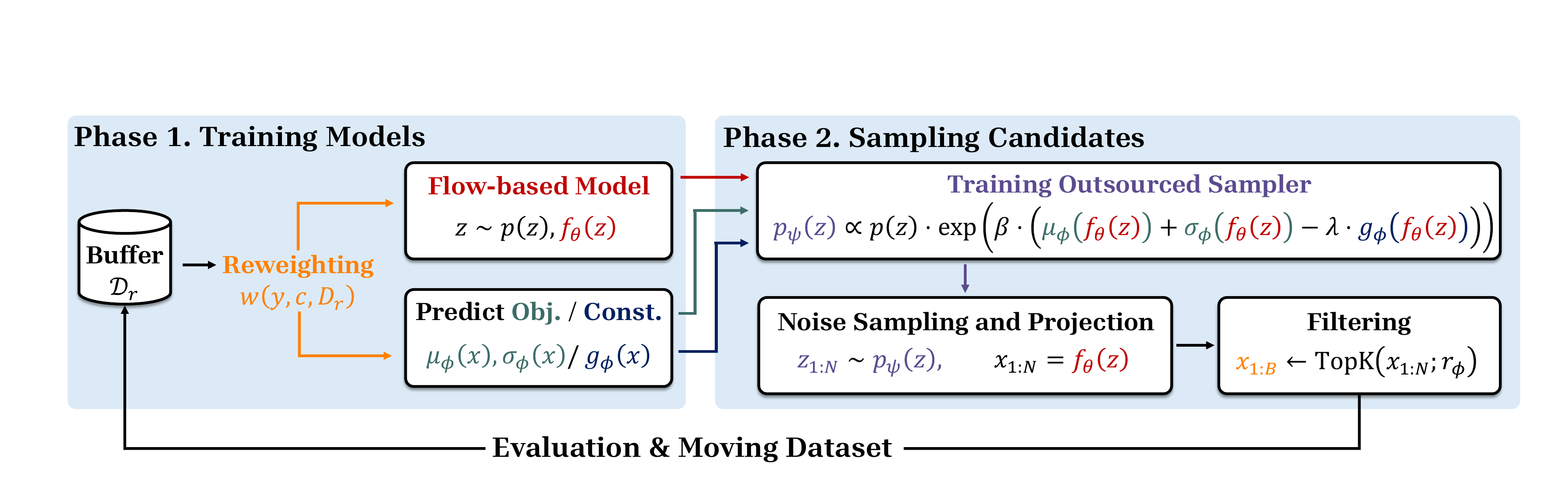}
    \vspace{-15pt}
    \caption{Overview of our method. \textbf{Phase 1:} Train flow-based models and proxies for the objective and constraints. \textbf{Phase 2:} Sample candidates from the posterior distribution using an outsourced diffusion sampler. After sampling, we utilize filtering to enhance sample efficiency. Then, we evaluate samples, update the dataset, and repeat the process until the evaluation budget is exhausted.}
    \label{fig:overview}
    \vspace{-12pt}
\end{figure}

To mitigate this issue, we adopt the outsourced diffusion sampler method proposed by~\citet{venkatraman2025outsourced}. In the outsourced diffusion sampler, we train an additional diffusion model that samples latents, which are then mapped into the data space via the deterministic mapping derived from the prior. Matching the distribution within the latent space significantly simplifies the alignment task when the distribution is highly multi-modal and has a large flat region in the original data space.

% Additionally, by fixing the original prior and configuring only the latent space, this method effectively mitigates mode collapse issues commonly encountered in other approaches.
% Meanwhile, naive implementations of reinforcement learning-based fine-tuning and adjoint-matching approaches can introduce biases and be prone to mode collapse \cite{venkatraman2024amortizing, venkatraman2025outsourced}.

% providing an unbiased posterior sampler. Although this method requires training an additional diffusion sampler in the fully noised latent space, the smoother distribution in this space significantly simplifies matching original highly multi-modal and large flat region target distributions.
\section{Preliminaries}
\subsection{Constrained Black-box Optimization}
% In constrained black-box optimization, our problem is:
We consider the following constrained black-box optimization problem.
\begin{align}
    \label{CBO}
    &\text{find }\mathbf{x}^*=\arg\max_{\mathbf{x} \in \mathcal{X}}f(\mathbf{x}) \quad\text{s.t. }\:g^{(1)}(\mathbf{x})\leq0,\cdots,g^{(M)}(\mathbf{x})\leq 0
    % &\text{with }R\text{ rounds of batch size }B\text{ queries}
\end{align}
The objective function \(f: \mathcal{X} \rightarrow \mathbb{R}\) and constraints \( g^{(1)},\cdots,g^{(M)}: \mathcal{X} \rightarrow \mathbb{R}\) are black-box functions. We assume an evaluation budget of
$R$ rounds, each with a batch of $B$ queries. % We also consider a more challenging scenario, only access to information on whether we violate constraints or not, i.e., \( h^{(m)}(\mathbf{x})=\mathbb{I}[g^{(m)}(\mathbf{x})>0]\). We refer to this as an indicator constraint.
% We also consider a more challenging scenario where only binary feasibility information is available, that is, whether each constraint is violated. We refer to this as an indicator constraint.
We also consider a more challenging scenario in which the only available feedback is a binary indicator of whether each constraint is violated. We refer to this as an indicator constraint.

\subsection{Flow-based Models}
Flow-based models~\citep{lipmanflow, liuflow, albergobuilding} are a class of generative models for approximating a target distribution \(q(\mathbf{x})\).
% Flow-based models are defined via the deterministic ordinary differential equation (ODE):
Flow-based models are defined via the deterministic ordinary differential equation (ODE).
\begin{equation}
    d\mathbf{x}_t = v_\theta(\mathbf{x}_t, t)\,dt \quad  v_\theta(\mathbf{x}_t, t): \mathbb{R}^d \times [0,1] \rightarrow \mathbb{R}^d
\end{equation}
where \(v_\theta\) is a parameterized velocity field.

For each given velocity field, the corresponding flow \(\psi_\theta(\mathbf{x}_0, t): \mathbb{R}^d \times [0,1] \rightarrow \mathbb{R}^d\) satisfies:
\begin{equation}
    \frac{d}{dt}\psi_\theta(\mathbf{x}_0, t) = v_\theta(\psi_\theta(\mathbf{x}_0, t), t),\quad \psi_\theta(\mathbf{x}_0,0)=\mathbf{x}_0.
\end{equation}
The velocity field \( v_\theta(\mathbf{x}_t, t) \) generates a continuous probability path \( p_t \) induced by the flow:
\begin{equation}
    \mathbf{x}_t = \psi_\theta(\mathbf{x}_0, t) \sim p_t,\quad \text{where}\quad \mathbf{x}_0 \sim p_0.
\end{equation}
\textbf{Training Flow-based Models.}
We use Flow Matching~\citep{lipmanflow} to learn the \( v_\theta \) that generates a path interpolating smoothly between an initial distribution \( p_0 = p \) and a target distribution \( p_1 = q \).

We employ the simplest linear interpolation path \(
\mathbf{x}_t = (1 - t)\mathbf{x}_0 + t\mathbf{x}_1,\) with derivative \(\frac{d\mathbf{x}_t}{dt} = \mathbf{x}_1 - \mathbf{x}_0
\). % The conditional Flow Matching loss is expressed as:
The conditional Flow Matching loss takes the following form.
\begin{align}\label{eq:fm}
    \mathcal{L}_{\text{CFM}}(\theta)=
     \mathbb{E}_{\substack{\mathbf{x}_0 \sim \mathcal{N}(0,I),\; \mathbf{x}_1 \sim q(\mathbf{x}),\; t \sim \text{Unif}(0,1)}}
    \left[\| v_\theta(\mathbf{x}_t, t) - (\mathbf{x}_1 - \mathbf{x}_0)\|^2_2\right]
\end{align}

\subsection{Posterior Inference in Flow-based and Diffusion Models}\label{sec:outsource}
Given a pretrained flow-based prior $p_{\theta}(\mathbf{x})$ and a reward function $r(\mathbf{x})$, we often need to sample from the posterior distribution $p_{\text{post}}(\mathbf{x}) \propto p_\theta(\mathbf{x}) r(\mathbf{x})$. However, direct sampling from this posterior is mostly intractable.

% In this section, we introduce outsourced diffusion sampling~\citep{venkatraman2025outsourced} to solve the aforementioned problem.
% In outsourced diffusion sampling, we interpret the sampling process of generative models into a noise generation $\mathbf{z} \sim p(\mathbf{z})$, followed by a deterministic transformation $\mathbf{x} = f_\theta(\mathbf{z})$, where $p(\mathbf{z})$ is standard normal and $f_\theta$ represents the learned mapping derived by prior.
We introduce outsourced diffusion sampling~\citep{venkatraman2025outsourced} to address this intractability.
In outsourced diffusion sampling, we decompose the sampling process of generative models into a noise generation step $\mathbf{z} \sim p(\mathbf{z})$, followed by a deterministic transformation $\mathbf{x} = f_\theta(\mathbf{z})$, where $p(\mathbf{z})$ is standard normal and $f_\theta$ represents the learned mapping derived from the prior.
% , i.e., $p_{\theta}(\mathbf{x})=p(f_{\theta}^{-1}(\mathbf{x}))$.
Under this formulation, by Proposition 3.1 of \citet{venkatraman2025outsourced}, we can sample from the posterior distribution by substituting noise generation as \(\mathbf{z} \sim p_{\text{post}}(\mathbf{z}) \propto p(\mathbf{z}) r(f_\theta(\mathbf{z}))\).
% To approximate the target distribution \(p_\psi(\mathbf{z}) \approx p_{\text{post}}(\mathbf{z})\), we can learn the parameters of diffusion sampler \(\psi\) with the trajectory balance (TB) objective \citep{malkin2022trajectory}:
To approximate the target distribution \(p_\psi(\mathbf{z}) \approx p_{\text{post}}(\mathbf{z})\), we learn the parameters of the diffusion sampler \(\psi\) by minimizing the TB objective \citep{malkin2022trajectory}.
\begin{align}
    \label{eq:diffusion}
    &\mathcal{L}_{\text{TB}}(\mathbf{z}_{0:1};\psi) = \left(\log  \frac{Z_\psi p(\mathbf{z}_0)\prod_{i=0}^{T-1} p_F(\mathbf{z}_{(i+1)\Delta t}|\mathbf{z}_{i \Delta t};\psi)}{p(\mathbf{z}_1)r(f_\theta(\mathbf{z}_1))\prod_{i=1}^Tp_B(\mathbf{z}_{(i-1)\Delta t}|\mathbf{z}_{i\Delta t})}\right)^2,
\end{align}

where \(Z_\psi\) is the parameterized partition function estimator, \((\mathbf{z}_0 \rightarrow \mathbf{z}_{\Delta_t}\rightarrow \cdots \mathbf{z}_1=\mathbf{z})\) is the discrete time Markov chain of reverse-time stochastic differential equation (SDE) with time increment \(\Delta t = \frac{1}{T}\).  \(p_F\) and \(p_B\) are transition kernels of the discretized reverse and forward SDE.

\section{Method}
% In this section, we introduce \textbf{CiBO}, a new framework for scalable constrained black-box optimization by leveraging generative models. Our method consists of two iterative stages. First, we train a flow-based model to capture the data distribution and surrogate models to predict objective values and constraints with uncertainty quantification. Next, we sample candidates from the posterior distribution. To accomplish this, we train a diffusion sampler that draws samples from the posterior distribution in the latent space. After sampling, we evaluate candidates, update the dataset, and repeat the process until the evaluation budget is exhausted. \Cref{fig:overview} illustrates the overview of our method.
% We introduce \textbf{CiBO}, a new framework for scalable constrained black-box optimization that leverages the generative model in candidate selection. Our method consists of two iterative stages. First, we train a flow-based model to capture the data distribution alongside surrogate models that predict objective values and constraints with uncertainty quantification. We then sample candidates by training a diffusion sampler that amortizes the posterior distribution in the latent space of the flow-based model. In other words, we first sample latents and project them into data space via the deterministic mapping from the flow-based model. After sampling, we evaluate candidates, update the dataset, and repeat until the evaluation budget is exhausted. \Cref{fig:overview} illustrates the overview of our method.
We introduce \textbf{CiBO}, a new framework for scalable constrained black-box optimization that leverages generative models for candidate selection. Our method consists of two iterative phases. First, we train a flow-based model to capture the data distribution alongside surrogate models that predict objective values and constraints. % We then train a diffusion sampler that amortizes the posterior distribution in the latent space of the flow-based model and draw candidates from it. Concretely, we first sample latents and project them into data space via its deterministic mapping.
We then train a diffusion sampler that amortizes the posterior distribution in the latent space of the flow-based model, and draw candidates by projecting sampled latents into data space via its deterministic mapping. After sampling, we evaluate candidates, update the dataset, and repeat until the evaluation budget is exhausted. \Cref{fig:overview} illustrates the overview of our method.

\subsection{Phase 1. Training Models}
\label{Method:Training Models}

% In each round $r$, we have a pre-collected dataset  $\mathcal{D}_{r}=\{\mathbf{x}_i, y_i, \mathbf{c}_i\}_{i=1}^{I}$, where $y_i=f(\mathbf{x}_i)$, $\mathbf{c}_i=(g^{(1)}(\mathbf{x}_i), \cdots, g^{(M)}(\mathbf{x}_i))$, and $I$ is the number of data points.
In each round $r$, we have a pre-collected dataset  $\mathcal{D}_{r}=\{\mathbf{x}_i, y_i, \mathbf{c}_i\}_{i=1}^{I}$, where $y_i=f(\mathbf{x}_i)$,  $\mathbf{c}_i=\{c_i^{m}\vert c_i^{m}=g^{(m)}(\mathbf{x}_i),\;\forall m=1,\dots,M\}$, and $I$ is the number of data points collected so far.

% \textbf{Training Prior.} We first train a prior model $p_{\theta}$ to capture the current data distribution. As the search space is too high-dimensional, it is better to implicitly constrain the search space close to the current data distribution. We use flow-based models to learn this distribution using \Cref{eq:fm}.
\textbf{Training Prior.} As the search space is high-dimensional, it is beneficial to implicitly constrain the search space close to the current data distribution. To this end, we first train a prior model $p_{\theta}$ to capture the current data distribution.

% \noindent\textbf{Training Surrogates.} We also train surrogate models to predict both objective values and constraints. As we are only able to access a small number of data points in the vast search space, we need to properly quantify the uncertainty of the prediction.
\noindent\textbf{Training Surrogates.} We also train surrogate models to predict both objective values and constraints. % Because we can access only a small number of data points in the vast search space, we need to properly quantify the uncertainty of their predictions. To this end, we train an ensemble of proxies to estimate objective values with uncertainty quantification \citep{lakshminarayanan2017simple}. Specifically, we train an ensemble of $K$ proxies $f_{\phi_1},\dots,f_{\phi_K}$ for objective values, and individual proxy $g_{\phi}^{(1)},\dots,g_{\phi}^{(M)}$ for each constraint.
Because we can access only a small number of data points in the vast search space, we need to properly quantify the uncertainty of their predictions. % To this end, we train an ensemble of $K$ proxies $f_{\phi_1},\dots,f_{\phi_K}$ for the objective and individual proxies $g_{\phi}^{(1)},\dots,g_{\phi}^{(M)}$ for each constraint, following the deep ensemble method of \citet{lakshminarayanan2017simple}.
To this end, we train an ensemble of $K$ proxies $f_{\phi_1},\dots,f_{\phi_K}$ for the objective, following the deep ensemble method of \citet{lakshminarayanan2017simple}. For the constraints, we train individual proxies $g_{\phi}^{(1)},\dots,g_{\phi}^{(M)}$ without ensembling, since using an ensemble for each of the $M$ constraints would lead to prohibitive time complexity.

\noindent\textbf{Reweighted Training.}
% high-data points -> promising data points
% During training, we introduce a reweighted training~\citep{tripp2020sample, kumar2020model, krishnamoorthy2023diffusion, kim2024bootstrapped} to focus on promising data points with high objective values while not violating constraints.
% Following ~\citet{tripp2020sample}, we assign a weight to each data point as follows:
We introduce reweighted training~\citep{tripp2020sample, kumar2020model, krishnamoorthy2023diffusion, kim2024bootstrapped} to focus on promising data points with high objective values while not violating constraints.
Following \citet{tripp2020sample}, we assign a weight to each data point:
\begin{equation}
    l(y,\mathbf{c}) =  y -\lambda \sum_{m=1}^{M} \max(0, c^m),\quad
    \label{eq:weighted sampler}
    w \left(y, \mathbf{c}, \mathcal {D}_r \right)=\frac{1}{\kappa|\mathcal{D}_r|+\text{rank}_{l, \mathcal{D}_r}(y,\mathbf{c})},
\end{equation}

% \begin{equation}
%     \label{eq:weighted sampler}
%     w \left(y, \mathbf{c}, \mathcal {D}_r \right)=\frac{\exp \left(l\left(y, \mathbf{c} \right)\right)}{\sum_{\left(y', \mathbf{c}' \right) \in \mathcal{D}_{r}}\exp \left(l\left(y', \mathbf{c}'\right)\right)}.
% \end{equation}

where $\lambda$ is a Lagrange multiplier, $\kappa \geq 0$ controls the strength of reweighting, and $\text{rank}_{l,\mathcal{D}_r}(y,\mathbf{c})$ is the rank of $(y,\mathbf{c})$ within $\mathcal{D}_r$ ordered by $l$.
% Then, our training objective for proxies and flow-based models can be described as follows:
We then define the training objective for the proxies and the flow-based model via reweighting as follows:
\begin{align}
    \label{eq:proxy}
    &\mathcal{L}(\phi) = \sum_{(\mathbf{x},y, \mathbf{c})\in\mathcal {D}_r}
     w(y, \mathbf{c},  \mathcal {D}_r) \cdot \left[ \sum_{k=1}^{K} \left( y-f_{\phi_{k}} \left(\mathbf{x}\right) \right)^2  +  \sum_{m=1}^{M} \left( c^m-g_{\phi}^{(m)} \left(\mathbf{x}\right) \right)^2 \right],
\end{align}
\begin{align}
    \label{eq:prior}
    &\mathcal{L}(\theta) = \mathbb{E}_{\mathbf{x}_0\sim \mathcal{N}(0,I) ,(\mathbf{x},y, \mathbf{c})\in\mathcal {D}_r,\,t \sim \text{Unif}(0,1) } \left[  w \| v_\theta(\mathbf{x}_t, t) - (\mathbf{x} - \mathbf{x}_0)\|^2_2\right].
\end{align}
\subsection{Phase 2. Sampling Candidates}
\label{Method:Sampling Candidates}
% After training models, we proceed to select candidates to evaluate in the next round.
After training, we select candidates for evaluation in the next round.
As the search space is high-dimensional, the prediction of surrogate models is likely to be inaccurate in regions that are too far away from the dataset collected so far. Therefore, it is advantageous to sample candidates from the distribution that satisfies the two desiderata: (1) promote exploration towards high-scoring and feasible regions, and (2) prevent sampling candidates that deviate too far from the current data distribution.
% To accomplish these objectives, we cast the candidate selection problem as sampling from the target distribution $p_{\text{post}}$ defined as follows:
To accomplish these objectives, we cast candidate selection as sampling from the following target distribution $p_{\text{post}}$.
% to efficiently explore the high-dimensional search space. Formally, we set our target distribution as:
\begin{equation}
\label{eq:KL Regularization}
    p_{\text{post}}(\mathbf{x}) = \arg\max_{p \in \mathcal{P}} \mathbb{E}_{\mathbf{x} \sim p}\left[ r_{\phi}(\mathbf{x}) \right] - \frac{1}\beta\cdot D_{\text{KL}}\left(p \,\|\, p_{\theta}\right),
\end{equation}
where $\mathcal{P}$ is the space of all probability distributions over the domain $\mathcal{X}$, and
\begin{equation}
\label{eq:Lagrangian Predictive Objective}
    r_{\phi}(\mathbf{x})=\mu_{\phi} (\mathbf{x})+\gamma\cdot \sigma_{\phi}(\mathbf{x}) -\lambda\sum_{m=1}^M \max (0, g_\phi^{(m)}(\mathbf{x})).
\end{equation}
$\mu_{\phi}(\mathbf{x})$ and $\sigma_{\phi}(\mathbf{x})$ represent the mean and standard deviation from the ensemble of surrogate models for the objective. \(\beta\) is an inverse temperature, and \(\lambda\) is a Lagrange multiplier.

% Based on the derivation from \citet{nair2020awac}, we can analytically derive our target distribution as:
Following the derivation in \citet{nair2020awac}, the target distribution admits the closed-form expression
\begin{equation}
    \label{eq: unnormalized posterior}
    p_\text{post}(\mathbf{x}) \propto p_\theta(\mathbf{x})\exp\left(\beta \cdot\left[ r_\phi(\mathbf{x})\right]\right).
\end{equation}
If we treat the flow-based model $p_{\theta}(\mathbf{x})$ as a prior and the exponential term $\exp(\beta\cdot [r_\phi(\mathbf{x})])$ as a reward $r(\mathbf{x})$, then our objective is to sample from the posterior distribution.
% In our framework, we cast the candidate selection problem as sampling from the posterior distribution to efficiently explore the high-dimensional search space.
% While most of the BO-based methods select candidates by searching for inputs that maximize the acquisition function, it is highly likely to be sub-optimal when the search space is high-dimensional \citep{ament2023unexpected}.

\noindent\textbf{Amortized Inference in Latent Space.}
\label{Method:Amortized Inference}
% However, directly sampling from this posterior is intractable~\citep{feng2025on}. Also, the target posterior is highly multi-modal and has a large plateau due to the constraint penalties in \Cref{eq:Lagrangian Predictive Objective}. This makes finetuning-based methods~\citep{venkatraman2024amortizing, fan2023dpok} susceptible to mode collapse.
% To this end, we utilize an amortized sampler in the latent space suggested by~\citet{venkatraman2025outsourced}.
% Unfortunately, directly sampling from the target posterior distribution is intractable due to the hierarchical structure in the generation process of flow-based models~\cite{feng2025on}. One can adopt directly fine-tuning a prior generative model to obtain an amortized sampler $p_{\psi}\approx p_{\text{post}}$ via reinforcement learning-based approaches \citep{venkatraman2024amortizing, fan2023dpok}.
% However, due to the constraints penalty term, we observe that the target distribution is highly multi-modal and has a large plateau. In such cases, directly fine-tuning pretrained generative models is likely to lead to mode collapse. To this end, we introduce an amortized sampler in the latent space suggested by Venkatraman et al \citep{venkatraman2025outsourced}.
\begin{figure}[t]
    \centering
    \includegraphics[width=1\textwidth]{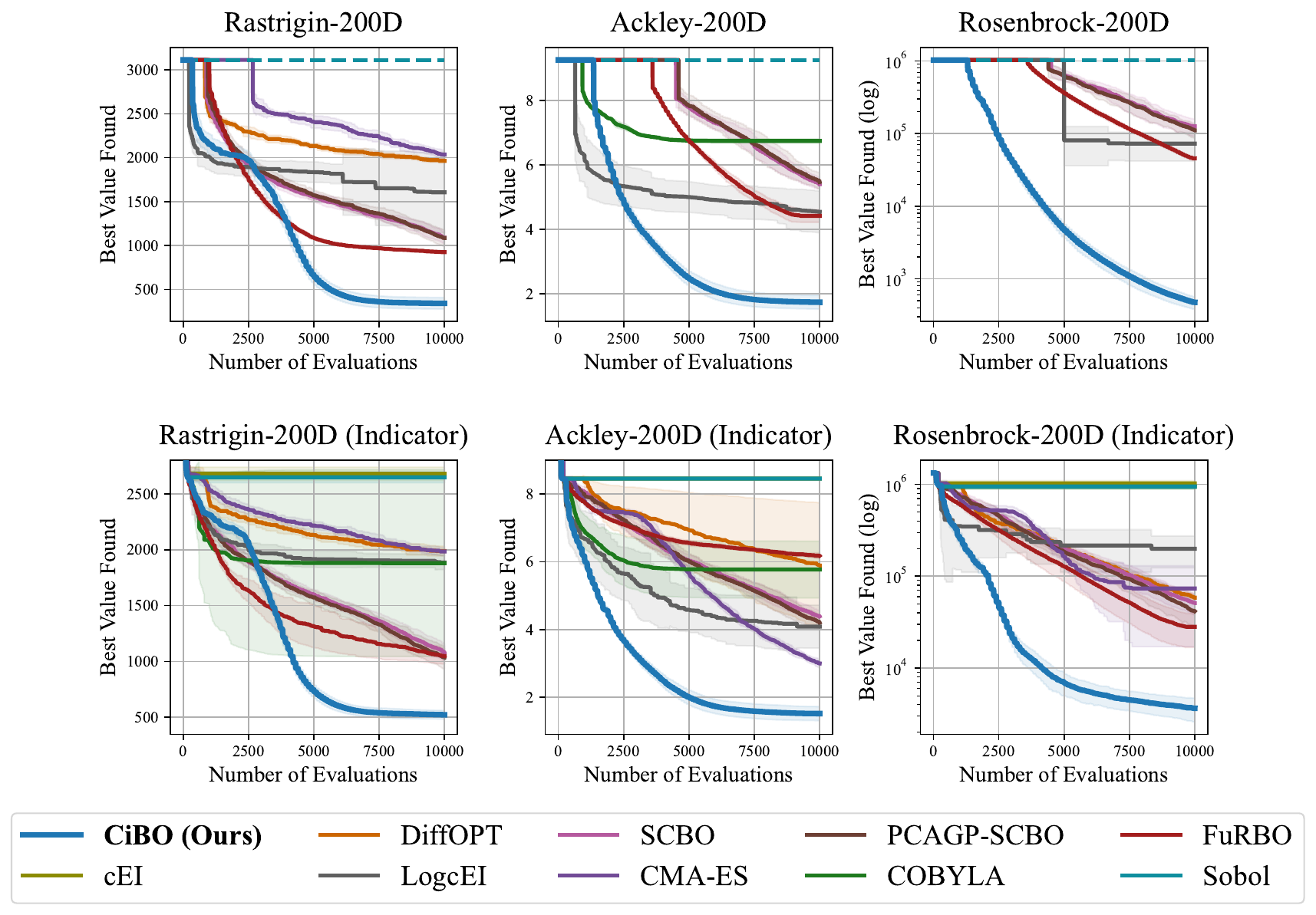}
    \vspace{-15pt}
    \caption {Comparison of CiBO with baselines on synthetic tasks. Experiments used 12 random seeds, reporting the mean and one standard deviation. A dashed line indicates no feasible solutions found.}
    \label{fig:synthetic}
    \vspace{-10pt}
\end{figure}
As introduced in \Cref{sec:outsource}, we can view the sampling procedure of flow-based models as drawing samples from the standard normal distribution $\mathbf{z} \sim p(\mathbf{z})$, followed by the deterministic transformation $\mathbf{x} =f_\theta(\mathbf{z})$, where $f_{\theta}$ is a deterministic mapping derived from the pretrained flow-based model.
% Within this framework, we can generate samples from the $p_{\text{post}}(\mathbf{x})$ by modifying the noise generation distribution as follows:
Within this framework, we can generate samples from $p_{\text{post}}(\mathbf{x})$ by modifying the latent generation distribution.
\begin{equation}
    \mathbf{z}\sim p_\text{post}(\mathbf{z})\propto p(\mathbf{z})r(f_\theta(\mathbf{z})).
\end{equation}
% To sample latents $\mathbf{z}$ from the posterior distribution in the latent space $p_\text{post}(\mathbf{z})$, we train a diffusion model $p_{\psi}(\mathbf{z})$ to amortize $p_\text{post}(\mathbf{z})$ with the following TB objective:
To sample latents $\mathbf{z}$ from the latent-space posterior $p_\text{post}(\mathbf{z})$, we train a diffusion model $p_{\psi}(\mathbf{z})$ to amortize $p_\text{post}(\mathbf{z})$ by minimizing the TB objective.
\begin{align}
    \label{eq:posterior}
    &\mathcal{L}_{\text{TB}}(\mathbf{z}_{0:1};\psi) = \left(\log  \frac{Z_\psi p(\mathbf{z}_0)\prod_{i=0}^{T-1} p_F(\mathbf{z}_{(i+1)\Delta t}|\mathbf{z}_{i \Delta t};\psi)}{p(\mathbf{z}_1)r(f_\theta(\mathbf{z}_1))\prod_{i=1}^Tp_B(\mathbf{z}_{(i-1)\Delta t}|\mathbf{z}_{i\Delta t})}\right)^2.
\end{align}
% We also adopt off-policy training to improve mode coverage, detailed in~Appendix~\ref{app: Diffusion Sampler Details}.

% By training an amortized sampler in the latent space of flow-based models, we can more accurately sample candidates from the target distribution as the posterior distribution in the latent space is smoother than that in the data space. Furthermore, since the TB objective supports off-policy training \citep{malkin2022trajectory}, it allows us to train the diffusion sampler using both on-policy trajectories from $p_F$ and off-policy trajectories from $p_B$. The latter are initialized from $\mathbf{z}_1$ collected from previous on-policy trajectories and stored in a replay buffer. This enhances mode coverage of the amortized sampler~\citep{sendera2024improved}.

\subsection{Filtering, Evaluation and Moving Dataset}
\noindent\textbf{Filtering.}
% After sampling from the posterior distribution, we need to carefully select candidates for the sample efficiency of the algorithm. To do so, we generate $N\cdot B$ samples from the amortized sampler and select the top-$B$ samples in terms of~\cref{eq:Lagrangian Predictive Objective} as querying candidates.
After sampling from the posterior distribution, we carefully select candidates to improve sample efficiency of our method. We generate $N\cdot B$ samples from the amortized sampler and select the top-$B$ samples based on \Cref{eq:Lagrangian Predictive Objective} as candidates.
% To this end, we introduce a filtering strategy. We generate $N\cdot B$ samples from the amortized sampler and select the top-$B$ samples in terms of Lagrangian relaxation of objectives as candidates.

\noindent\textbf{Evaluation and Dataset Management.}
% We evaluate the values of the objective function and constraint functions for each selected candidate.
We evaluate the objective function and constraint functions for each selected candidate. Then, we update the dataset with new observations. We empirically find that taking only a subset of total observations is beneficial in terms of both time complexity and sample efficiency. Therefore, we remove the samples with the lowest score if the size of the dataset is larger than the buffer size $L$.
% In other words, we update the dataset with new observations and remove the observations whose objective values are the lowest.
We present the pseudocode of our method in Appendix~\ref{app:algorithms}.

\section{Experiments}
% In this section, we report experimental results for scalable constrained black-box optimization tasks. First, we perform experiments on three 200-dimensional synthetic functions, which are the standard benchmarks in Bayesian Optimization (BO) literature~\citep{eriksson2019scalable}.
% Furthermore, we assess the performance of our method on a more challenging scenario, where the feedback from constraints is given as binary indicators of feasibility. We refer to this setting as the indicator constraint setting.
% Finally, we conduct experiments on four real-world optimization tasks: Rover Planning 60D~\citep{wang2018batched, eriksson2021scalable}, HalfCheetah 102D~\citep{todorov2012mujoco}, Mopta 124D~\citep{anjos2009mopta}, and Lasso DNA 180D~\citep{vsehic2022lassobench}. For all tasks, we report the best value found so far, and assign the largest found in all algorithms to the infeasible solutions, following \citet{hern2016general, eriksson2021scalable}. The detailed description of each task can be found in~Appendix~\ref{app:task-details}.
% We evaluate CiBO on three 200-dimensional synthetic functions and four real-world benchmarks spanning 60 to 180 dimensions. For the synthetic tasks, we also consider a more challenging indicator constraint setting where only binary feasibility feedback is available. For all tasks, we report the best feasible value found and assign the worst observed value across all methods to infeasible runs, following \citet{hern2016general, eriksson2021scalable}. Task details are in Appendix~\ref{app:task-details}.
We evaluate CiBO on three 200-dimensional synthetic functions and four real-world benchmarks spanning 60 to 180 dimensions. For the synthetic tasks, we also consider a more challenging indicator constraint setting where only binary feasibility feedback is available. For all tasks, we report the best feasible value found and assign the worst observed value across all methods to infeasible runs, following \citet{hern2016general, eriksson2021scalable}.
\begin{figure}[t]
    \centering
    \includegraphics[width=1\textwidth]{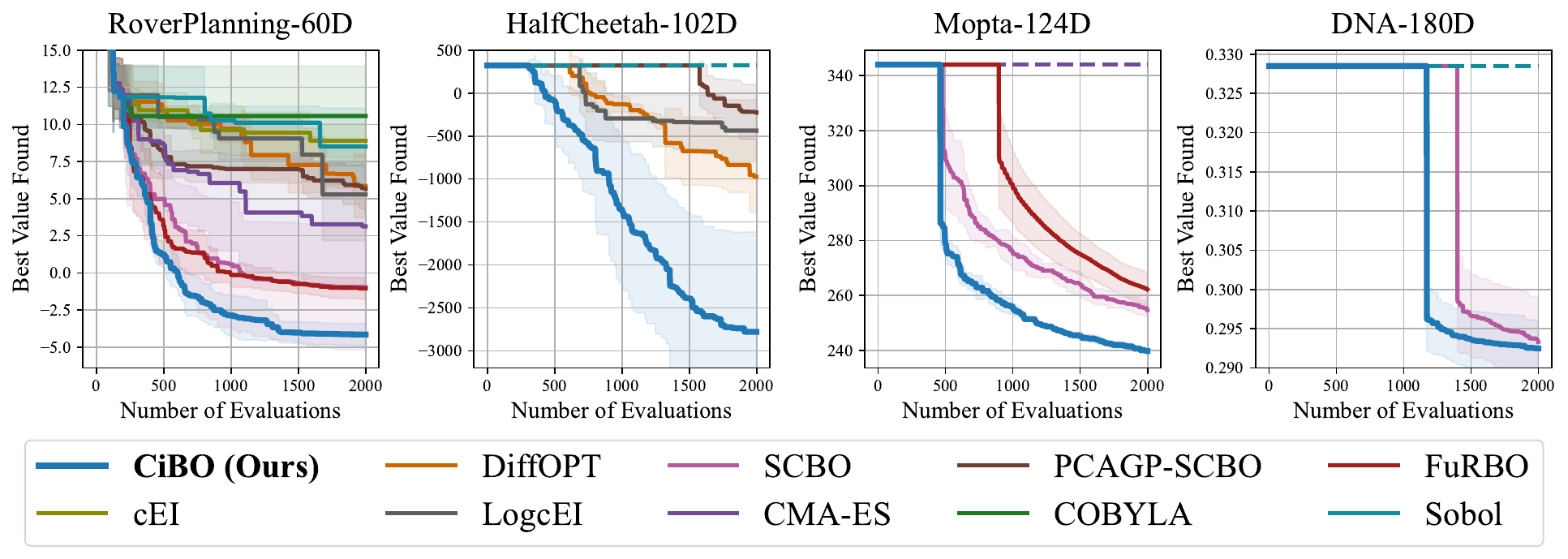}
    \caption {Comparison of CiBO with baselines on real world tasks. Experiments used 4 random seeds, reporting mean and one standard deviation. A dashed line indicates no feasible solutions found.}
    \label{fig:realworld}
    \vspace{-15pt}
\end{figure}

% \subsection{Tasks}
% For synthetic benchmarks, we evaluate on Rastrigin-200D, Ackley-200D, and Rosenbrock-200D, each subject to two standard inequality constraints from SCBO~\citep{eriksson2021scalable}: $\sum_{d=1}^{200} x_d \leq 0$ and $\|\mathbf{x}\|_2^2 \leq 30$. We use $|\mathcal{D}_0|=200$, batch size $B=100$, and a budget of $10{,}000$ evaluations. For the indicator constraint setting, we seed the initial dataset with 10 feasible points obtained via hit-and-run MCMC, as finding an initial feasible solution is otherwise too challenging for all baselines.
% For real-world benchmarks, we evaluate on Rover Planning 60D with 15 square-shaped obstacles, HalfCheetah 102D with 5 constraints, Mopta 124D with 68 constraints, and Lasso DNA 180D with 5 constraints. We initialize with $|\mathcal{D}_0|=200$ and a budget of $2{,}000$. We use batch size $B=50$ for all tasks except Mopta, where we use $B=20$, as baselines could not identify feasible solutions with $50$. Task details are in Appendix~\ref{app:task-details}.
\subsection{Tasks}
We consider three synthetic and four real-world benchmarks for evaluating our method.
\begin{itemize}[leftmargin=1em]
    \item \textbf{Synthetic.} Rastrigin ($D=200$), Ackley ($D=200$), and Rosenbrock ($D=200$), each subject to two inequality constraints: $\sum_{d=1}^{200} x_d \leq 0$ and $\|\mathbf{x}\|_2^2 \leq 30$. We also consider a more challenging indicator constraint setting where only binary feasibility feedback is available.
    \item \textbf{Real-world.} RoverPlanning ($D=60$, 15 constraints), HalfCheetah ($D=102$, 5 constraints), Mopta ($D=124$, 68 constraints), and Lasso DNA ($D=180$, 5 constraints).
\end{itemize}
For synthetic tasks, we use $|\mathcal{D}_0|=200$, batch size $B=100$, and a budget of $10{,}000$ evaluations. For the indicator constraint setting, we seed the initial dataset with 10 feasible points obtained via hit-and-run MCMC, as finding an initial feasible solution is otherwise too challenging for all baselines. For real-world tasks, we use $|\mathcal{D}_0|=200$, a budget of $2{,}000$ evaluations, and batch size $B=50$ for all tasks except Mopta ($B=20$), as baselines could not find feasible solutions with $B=50$. Please refer to Appendix~\ref{app:task-details} for detailed task descriptions.

\subsection{Baselines}
% We compare our method with several constrained BO baselines, including cEI~\citep{schonlau1998global},
% LogcEI~\citep{ament2023unexpected},
% SCBO~\citep{eriksson2021scalable},
% PCAGP-SCBO~\citep{maathuis2024high},
% FuRBO~\citep{ascia2025feasibility}, the evolutionary search algorithm
% CMA-ES~\citep{hansen2006cma}, the derivative-free optimizer COBYLA~\citep{powell1994direct}, and the quasi-random Sobol sequences~\citep{sobol1967distribution}.
% We also evaluate DiffOPT~\citep{kong2025diffusion}, a generative model-based approach designed for constrained optimization.
% Detailed settings of all baselines are provided in~Appendix~\ref{app:baselines-details}.
We compare with constrained expected improvement methods cEI~\citep{schonlau1998global} and LogcEI~\citep{ament2023unexpected}, which extend standard BO acquisition functions with constraint handling. We also evaluate SCBO~\citep{eriksson2021scalable}, PCAGP-SCBO~\citep{maathuis2024high}, and FuRBO~\citep{ascia2025feasibility}, which are specifically designed for scalable constrained black-box optimization. We further compare with DiffOPT~\citep{kong2025diffusion}, a generative model-based approach designed for constrained optimization. We also include the evolutionary search algorithm CMA-ES~\citep{hansen2006cma}, the derivative-free optimizer COBYLA~\citep{powell1994direct}, and the quasi-random Sobol sequences~\citep{sobol1967distribution}. Detailed settings are provided in Appendix~\ref{app:baselines-details}.

\subsection{Implementation}
We parametrize the surrogate models, the flow-based model, and the diffusion sampler all with lightweight MLPs. At every optimization round, we retrain the proxies for 100 steps, the flow model for 500 steps, and the diffusion sampler for 100 steps. As shown in the runtime breakdown in Appendix~\ref{app:Runtimes}, this lightweight design keeps the per-round computational cost comparable to or lower than that of BO-based approaches. Please refer to Appendix~\ref{app:implementation-details} for implementation details.

\subsection{Synthetic Experiments}
As shown in \Cref{fig:synthetic}, CiBO outperforms all baselines on all three tasks under both constraint settings. Among BO baselines, SCBO, PCAGP-SCBO, LogcEI, and FuRBO identify feasible solutions but plateau early due to the limited expressiveness of GP surrogates in 200 dimensions. cEI, CMA-ES, and COBYLA fail to find feasible solutions on some tasks, and DiffOPT struggles with low sample efficiency across several tasks. These results show that posterior inference in latent space enables CiBO to find feasible solutions and maximize the objective in a sample-efficient manner.

% balances the value optimization and feasibility, thereby improving sample efficiency.

% We introduce the Lagrangian Multiplier $\lambda$ for the inequality constraints. By incorporating the constraints into the objective, we can rewrite our objective as
% $\arg\max_{\mathbf{x} \in \mathcal{X}}f(\mathbf{x}) - \lambda \sum_j g_j(\mathbf{x})$.
% This formulation penalizes the objective value, so simply increasing the objective alone does not suffice.

% \textbf{Indicator Constraints}.
\begin{figure}[t]
  \centering
  \begin{subfigure}[t]{0.5\textwidth}
    \centering
    \includegraphics[width=0.95\linewidth,
                     ]{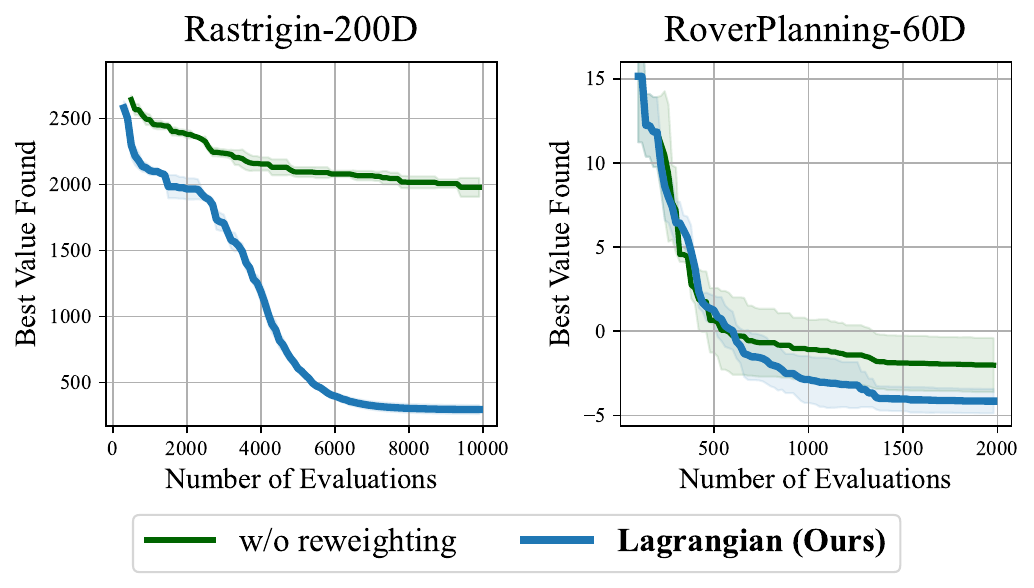}
    \caption{Reweighted Training}
    \label{fig:abl-a}
  \end{subfigure}\hfill
  \begin{subfigure}[t]{0.5\textwidth}
    \centering
    \includegraphics[width=0.95\linewidth,
                     ]{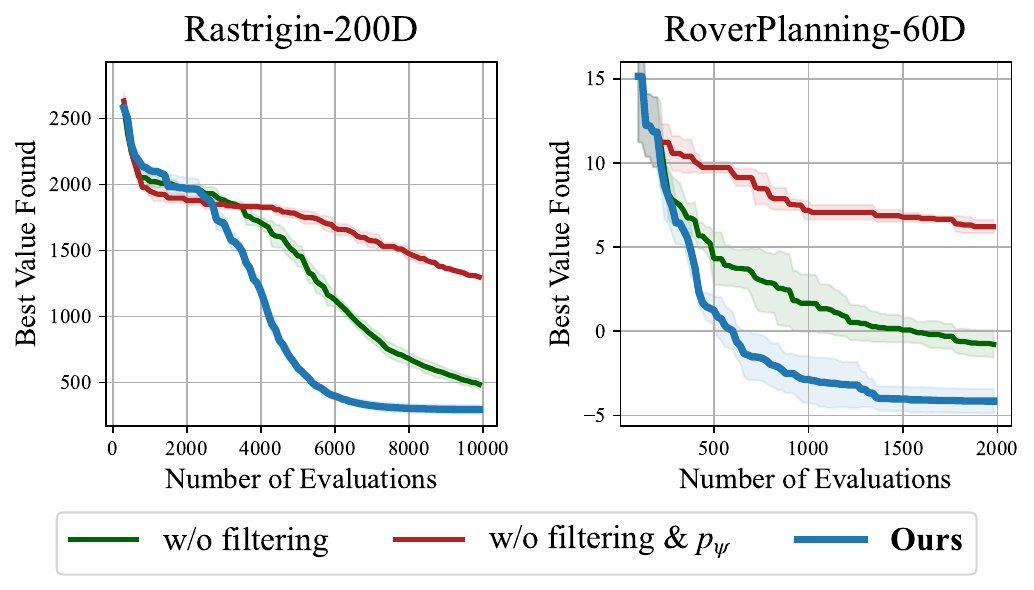}
    \caption{Sampling Procedure}
    \label{fig:abl-b}
  \end{subfigure}

  \vspace{.5em}

  \begin{subfigure}[t]{0.5\textwidth}
    \centering
    \includegraphics[width=0.95\linewidth,
                     ]{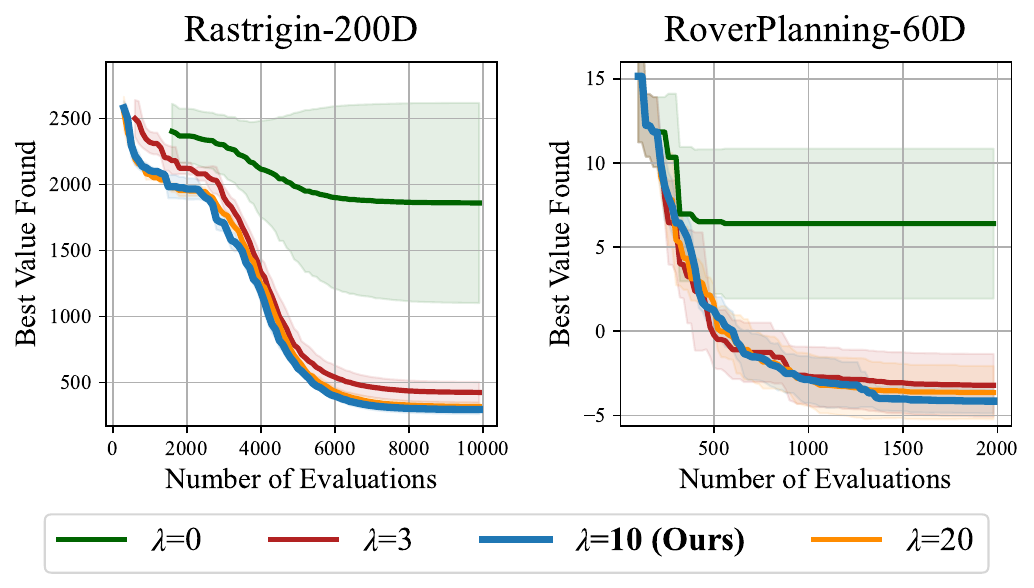}
    \caption{Analysis on $\lambda$}
    \label{fig:abl-c}
  \end{subfigure}\hfill
  \begin{subfigure}[t]{0.5\textwidth}
    \centering
    \includegraphics[width=0.95\linewidth,
                     ]{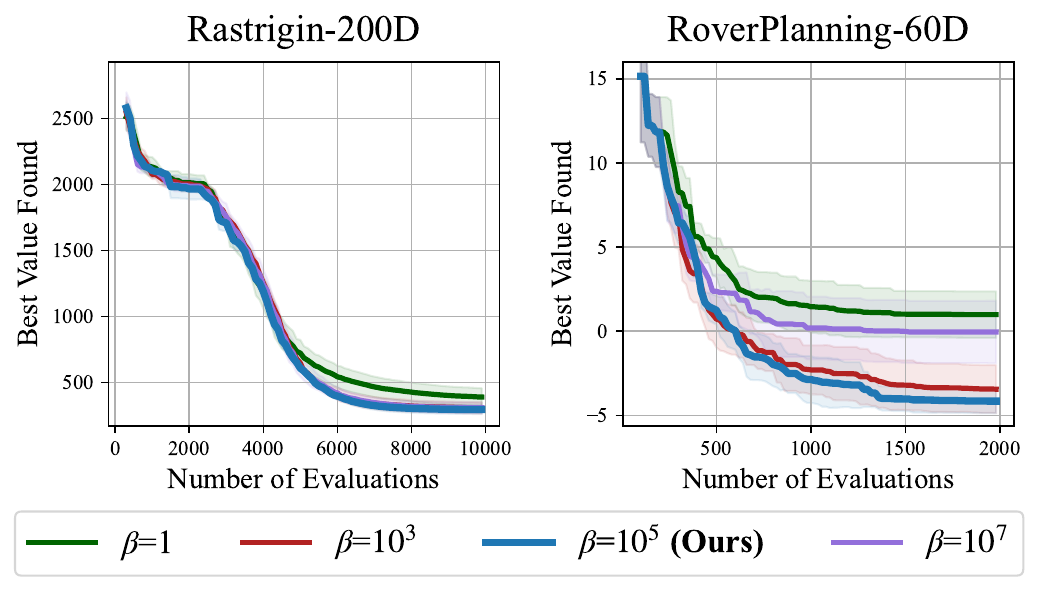}
    \caption{Analysis on $\beta$}
    \label{fig:abl-d}
  \end{subfigure}
  \vspace{-5pt}
  \caption{Additional analysis for various components of CiBO. Experiments are conducted with four random seeds, and the mean and one standard deviation are reported.}
  \label{fig:ablation}
    \vspace{-10pt}
\end{figure}

\subsection{Real World Experiments}
% We evaluate on four real-world benchmarks: Rover Planning 60D with 15 square-shaped obstacles, HalfCheetah 102D with 5 constraints, Mopta 124D with 68 constraints, and Lasso DNA 180D with 5 constraints. We initialize with $|\mathcal{D}_0|=200$ and a budget of $2{,}000$. We use batch size $B=50$ for all tasks except Mopta, where we use $B=20$, as baselines could not identify feasible solutions with $50$.

% As illustrated in \Cref{fig:realworld}, our approach consistently identifies high-quality feasible solutions with superior sample efficiency across all tasks. We observe that the performance gap between our method and competing baselines becomes substantially larger on real-world problems than on synthetic problems. Remarkably, most baselines failed to find any feasible solutions for the challenging Mopta-124D and DNA-180D tasks. While SCBO is the only competing method to achieve feasibility alongside our approach, it exhibits low sample efficiency.
% We also find that several generative model-based approaches struggle with identifying feasible solutions in real-world tasks with a large number of constraints.
% This again highlights the robustness of our approach to scalable constrained black-box optimization.
As shown in \Cref{fig:realworld}, CiBO consistently identifies high-quality feasible solutions with superior sample efficiency across all tasks, and the performance gap relative to competing baselines is substantially more pronounced than in the synthetic tasks. DiffOPT achieves feasibility only on HalfCheetah, failing to identify feasible solutions on Mopta-124D and DNA-180D. FuRBO recovers feasible solutions on Mopta, which involves a large number of constraints, yet exhibits limited overall performance. SCBO is the sole competing method to achieve feasibility across most tasks but with low sample efficiency, further highlighting the robustness of CiBO in scalable constrained settings.

\subsection{Additional Analysis}
\label{sec:additional analysis}
In this section, we conduct an analysis of each component of our proposed method through ablation experiments on Rastrigin 200D and Rover Planning 60D tasks.

\paragraph{Feasibility Ratio.}
We also conduct feasibility-ratio analysis of the proposed candidates of CiBO. As depicted in Appendix~\ref{app:feasibility}, CiBO reaches near-perfect feasibility within 5$\sim$10 batches on Rastrigin-200D, while BO baselines require roughly twice as many batches to reach comparable ratios. On RoverPlanning-60D, most baselines remain at a low feasibility ratio, whereas CiBO maintains steady improvement throughout the optimization.

\paragraph{Reweighted Training.}
To investigate the effectiveness of the reweighted training approach in \Cref{eq:weighted sampler}, we compare against a variant trained without reweighting (i.e., uniform sampling). As shown in \Cref{fig:abl-a}, removing reweighting significantly reduces sample efficiency, indicating that it effectively handles objectives and feasibility in high-dimensional space. We further compare different reweighting strategies (e.g., score-weighted vs.\ rank-based) in Appendix~\ref{app:reweighting}.

\paragraph{Sampling Procedure.}
To analyze the effect of each component during candidate sampling, we experiment with two variants: removing filtering, and removing both filtering and the diffusion sampler, thus sampling candidates directly from the prior $p_{\theta}$.
As shown in \Cref{fig:abl-b}, a significant gap exists between ours and the other variants, validating the effectiveness of each proposed component. We also employ off-policy training for the amortized diffusion sampler to improve mode coverage of the posterior distribution. An ablation on this component is provided in Appendix~\ref{app:policy}.

\paragraph{Lagrange Multiplier $\lambda$.}
% We introduce the Lagrangian multiplier $\lambda$ to balance between objective values and constraint violations.
% To assess the impact of penalty strength, we experiment with different values of the Lagrange multiplier $\lambda$. As shown in \Cref{fig:abl-c}, setting $\lambda = 0$ (eliminating the Lagrangian penalties) significantly degrades performance on both tasks, as it only focuses on high objective values and neglects the feasibility of solutions.
% Conversely, when $\lambda>0$, it shows robustness to the choice of $\lambda$ and improves performance significantly.
To assess the impact of penalty strength, we experiment with different values of the $\lambda$.
As shown in \Cref{fig:abl-c}, setting $\lambda=0$ significantly degrades performance on both tasks, as the method then neglects feasibility. Conversely, our method shows robustness to the specific choice of $\lambda > 0$, consistently improving performance. We also compare against an adaptive $\lambda$ schedule in Appendix~\ref{app:adaptive-lambda} and find that it adds complexity without improving performance.

\paragraph{Inverse Temperature $\beta$.}
% The inverse temperature controls the balance between the prior $p_\theta(\mathbf{x})$ and the reward function $r(\mathbf{x})$.
% We conduct experiments by varying $\beta$ values. As shown in \Cref{fig:abl-d}, using a moderately high $\beta$ generally helps to improve sample efficiency.
% However, if $\beta$ is too high, the performance is heavily dependent on the accuracy of surrogate models, leading to sub-optimal results. Conversely, too small $\beta$ does not move the posterior distribution towards high-scoring regions, thereby leading to slow convergence.
The inverse temperature controls the balance between the prior $p_\theta(\mathbf{x})$ and the reward function $r(\mathbf{x})$.
While we fix $\beta=10^5$ across all experiments, we conduct ablation studies on the controllability of $\beta$ in \Cref{fig:abl-d}. A moderately high $\beta$ improves sample efficiency, but if $\beta$ is too high, performance becomes sensitive to surrogate model accuracy, leading to sub-optimal results. Too small $\beta$ leads to slow convergence due to lack of guidance toward high-scoring regions.

\paragraph{Buffer size $L$.}
For each round, we maintain a subset of the collected dataset with a buffer size $L$. We set $L=2000$ for synthetic tasks and $L=1000$ for real-world tasks, as the evaluation budget for real-world tasks is 2000. We analyze the effect of $L$ in Appendix~\ref{app:buffer} and find that using a small $L$ leads to early saturation and sub-optimal solutions, while a large $L$ slows the rate of improvement.

\paragraph{Initial Dataset size $|\mathcal{D}_0|$ and Batch size $B$.}
% Experiment configurations $\vert\mathcal{D}_0\vert$ and $B$ can be critical for the performance of various methods. We conduct analysis in Appendix~\ref{app:batch} and find that our method maintains consistent performance across different initial experiment configurations.
The initial dataset size $|\mathcal{D}_0|$ and batch size $B$ are important design choices that can significantly affect the performance of optimization methods. A small $|\mathcal{D}_0|$ may provide insufficient coverage of the search space, while a large $B$ increases the per-round evaluation cost. We analyze sensitivity to both in Appendix~\ref{app:batch} and find that CiBO maintains consistent performance across different configurations, demonstrating robustness to these choices.

\paragraph{Misaligned Constraints.}
In the synthetic tasks, the optimal points of Rastrigin and Ackley coincide with the center of the norm constraint, which may artificially favor all methods. To verify that CiBO does not rely on this alignment, we test on both Rastrigin-200D and Ackley-200D with shifted constraint centers. As shown in Appendix~\ref{app:misaligned}, CiBO consistently outperforms the baselines on both tasks, confirming that its advantage stems from expressive posterior inference rather than constraint-objective alignment.

\paragraph{Comparison with Unconstrained Generative Methods.}
DDOM~\citep{krishnamoorthy2023diffusion} and DiBO~\citep{yun2025posterior} are generative model-based methods designed for unconstrained optimization. We adapt both methods to the constrained setting via Lagrangian relaxation with independently searched $\lambda$ and provide a comparison in Appendix~\ref{app:unconstrained}. As shown in the results, both methods underperform CiBO, highlighting the importance of explicitly incorporating constraint handling into the framework.

\paragraph{Scalability.} We report a per-round runtime breakdown in Appendix~\ref{app:Runtimes}, decomposing each round into surrogate fitting, generative model training, candidate proposal, and oracle evaluation. Even with several components to train, CiBO exhibits a low runtime per round as it uses lightweight models for both surrogate and generative stages, avoiding the expensive GP fitting in BO-based methods.

\section{Conclusion}
\label{sec:conclusion}
We introduced CiBO, a generative model-based framework for scalable constrained black-box optimization. Our approach formulates candidate selection as posterior inference, leveraging flow-based models to capture the data distribution and surrogate models to predict both objectives and constraints.
By amortizing posterior sampling in the latent space with outsourced diffusion samplers, our method effectively addresses the challenges posed by highly multi-modal and flat posterior distributions that arise from incorporating constraints.
% mitigating issues such as mode collapse and poor scalability that affect prior methods.
Extensive experiments across synthetic and real-world benchmarks demonstrate the superiority of our proposed method.

\paragraph{Limitations and Future Works.}
% \section{Limitations}
% \noindent\textbf{Limitations and Future work.}
We are interested in improving our method.
First, as we need to train all models with the updated dataset in every round, presenting a framework that can reuse the trained models from the previous rounds would be beneficial. Furthermore, there are several advancements in the literature on flow-based model training \citep{tong2024improving} and diffusion samplers \citep{havens2025adjoint, kim2025adaptive}, which could potentially yield substantial performance gains. We leave them as future work.

\clearpage
\section*{Impact Statement}
Advances in real-world design optimization have the potential to drive major innovations, but they also come with potential risks and unintended consequences. For example, optimization techniques in biochemical design may uncover novel compounds with therapeutic potential, but similar methods could also be misused to discover harmful substances. It is essential for researchers to act responsibly and ensure their work serves the public good.

\bibliographystyle{unsrtnat}
\bibliography{example_paper}

%%%%%%%%%%%%%%%%%%%%%%%%%%%%%%%%%%%%%%%%%%%%%%%%%%%%%%%%%%%%%%%%%%%%%%%%%%%%%%%
%%%%%%%%%%%%%%%%%%%%%%%%%%%%%%%%%%%%%%%%%%%%%%%%%%%%%%%%%%%%%%%%%%%%%%%%%%%%%%%
% APPENDIX
%%%%%%%%%%%%%%%%%%%%%%%%%%%%%%%%%%%%%%%%%%%%%%%%%%%%%%%%%%%%%%%%%%%%%%%%%%%%%%%
%%%%%%%%%%%%%%%%%%%%%%%%%%%%%%%%%%%%%%%%%%%%%%%%%%%%%%%%%%%%%%%%%%%%%%%%%%%%%%%
\newpage
\appendix
\section*{Appendix}

\section{Algorithms}
\label{app:algorithms}
\begin{algorithm}[]
\caption{CiBO}
\label{alg1}
\begin{algorithmic}[1]
    \STATE \textbf{Input:} 
        Initial dataset \(\mathcal{D}_0\);
        Max rounds \(R\);
        Batch size \(B\); 
        Buffer size \(L\); Number of constraints \(M\); \\
        Flow model $p_{\theta}$;
        Diffusion sampler $p_{\psi}$;
        Proxies $f_{\phi_1}, \cdots, f_{\phi_K}, g_{\phi}^{(1)}, \cdots ,g_{\phi}^{(M)} $;
    \FOR{\(r = 0, \ldots, R-1\)}
        \STATE Initialize $p_\theta, p_\psi$, $f_{\phi_1}, \cdots, f_{\phi_K},g_{\phi}^{(1)}, \cdots ,g_{\phi}^{(M)}$
        \STATE
        \STATE \textbf{Phase 1. Training Models}
        \STATE Compute weights $w(y, \mathbf{c}, \mathcal{D}_r)$ with \Cref{eq:weighted sampler}
        \STATE Train $p_{\theta}$ with \Cref{eq:prior}
        \STATE Train  $f_{\phi_1}, \cdots, f_{\phi_K},g_{\phi}^{(1)}, \cdots ,g_{\phi}^{(M)}$ with \Cref{eq:proxy}
        \STATE
        
        \STATE \textbf{Phase 2. Sampling Candidates}
        \STATE Train  $p_\psi$ with \Cref{eq:posterior} using prior $p_{\theta}$ and $\mathbf{z} \sim N(0,\mathbf{I})$  
        
        \STATE Sample latent noise with \(\{\mathbf{z}_i\}_{i=1}^{N B} \sim p_\psi(\mathbf{z})\)
        \STATE Projection to data space with learned mapping \(\mathbf{x}_{i} = f_\theta(\mathbf{z}_i) \quad \forall i \in \{1, \cdots, NB\}\)
        % \STATE Filter top-$B$ samples 
        % \(\{\mathbf{x}_b\}_{b=1}^B\) among 
        % \(\{\mathbf{x}_i\}_{i=1}^N\) with $r_\phi(\mathbf{x}) - \lambda \sum_{m=1}^M g_\phi^{(m)}(\mathbf{x})$
        \STATE
        \STATE \textbf{Filtering} 
        \STATE Select top-$B$ samples $\{\mathbf{x}_b\}_{b=1}^B$ with respect to $r_\phi(\mathbf{x})$ 
        % \(\{\mathbf{x}_b\}_{b=1}^B\) with respect to: \\$r_\phi(\mathbf{x}_i) - \lambda \sum_{m=1}^M \max (0, g_\phi^{(m)}(\mathbf{x}_i)) \quad \forall i= \{1,\cdots,NB\}$
        % \FOR{j=0, ..., J-1}
        \STATE
        \STATE \textbf{Evaluation and Moving Dataset}
        \STATE Evaluate $y_b=f(\mathbf{x}_b),\quad c^m_b=g^{(m)}(\mathbf{x}_b)\quad\forall m=\{1,\cdots M\}\quad\forall b= \{1,\cdots,B\}$
        \STATE Update \(\mathcal{D}_{r+1} \leftarrow\mathcal{D}_r\cup\{(\mathbf{x}_b, y_b, \mathbf{c}_b)\}_{b=1}^B\)
        
        \IF{$\vert\mathcal{D}_{r+1}\vert>L$}
            \STATE Remove last
            $ \vert\mathcal{D}_{r+1}\vert-L$ samples from $\mathcal{D}_{r+1}$ with respect to: $y - \lambda\sum_{m=1}^M\text{max}(0,c^m)$
        \ENDIF

    \ENDFOR
\end{algorithmic}

\end{algorithm}

\clearpage

\section{Task Details}
\label{app:task-details}

\subsection{Synthetic Functions}
\label{app:synthetic details}
We evaluate three synthetic functions in our constrained black-box optimization experiments: Rastrigin, Ackley, and Rosenbrock. The Rastrigin and Ackley functions are highly multi-modal functions with numerous local minima, whereas the Rosenbrock function features a narrow valley that makes convergence to the global minimum notoriously difficult \citep{demo2021supervised}. Following \cite{wang2020learning, yi2024improving}, we define the search domains as Rastrigin: \([-5,5]^D\), Ackley: \( [-5, 10]^D\), and Rosenbrock: $[-5, 10]^D$. All functions are subject to two constraints:
$$\sum_{d=1}^{200} {x}_d \leq 0 \quad \text{and} \quad ||\mathbf{x}||_2^2 \leq 30$$
Although prior work enforced the tighter bound $||\mathbf{x}||_2^2 \leq 5$, we relax this constraint in our high-dimensional setting.
% We also conducted experiments using these constraints in indicator form:
% $$ \mathbb{I}[10 - \sum_i^D \mathbf{x}_i >0] \quad \text{and}  \quad \mathbb{I}[30 - ||\mathbf{x}||_2^2 >0]$$
For the indicator constraint experiments, we sample initial feasible points by hit-and-run MCMC \citep{zabinsky2013hit}.

\subsection{Rover Trajectory Planning}
 Rover Trajectory Planning is a trajectory optimization task in a 2D environment introduced by \citet{wang2018batched}. The objective is to optimize the rover's trajectory, where its trajectory is represented by 30 points defining a B-Spline. We place 15 impassable obstacles $o_i$ and impose collision-avoidance constraints $c_i(\mathbf{x})$ as in \citet{eriksson2021scalable}:
 $$c_i(\mathbf{x}) =
\begin{cases}
  -\,d\bigl(o_i,\gamma(\mathbf{x})\bigr)
    & \text{if }\gamma(\mathbf{x})\cap o_i = \varnothing, \\[6pt]
  \displaystyle
  \max_{\alpha \in \gamma(\mathbf{x})\cap o_i}
  \;\min_{\beta \in \partial o_i}
    d(\alpha,\beta)
    & \text{otherwise.}
\end{cases}$$
where $\gamma(\mathbf{x})$ denotes final trajectory, $o_i$ is the region of the obstacle and $\partial o_i$ denotes the boundary of $o_i$. A trajectory is feasible if and only if it does not intersect any obstacle. We follow the implementation from \citet{wang2018batched}, but since there is no released code for the constraints, we implement the violation metric ourselves. Below is an example of the trajectory found by our method.

\begin{figure}[!h]
    \centering
    \includegraphics[scale=0.5]{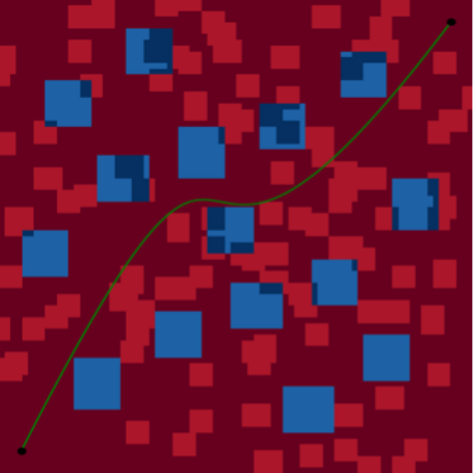}
    \caption {Trajectory found by CiBO, achieving minimum value of -4.14.}
    \label{fig:example of rover planning}
    \vspace{-10pt}
\end{figure}

\subsection{MuJoCo locomotion}
The MuJoCo locomotion task \citep{todorov2012mujoco} is a widely used benchmark in Reinforcement Learning. In this context, we aim to optimize a linear policy represented by the equation \(\mathbf{a} = \mathbf{W}\mathbf{s}\), where \(\mathbf{a}\) is the action and \(\mathbf{s}\) is the state. Our objective is to maximize the average return of this policy by identifying the optimal weight matrix \(\mathbf{W}\).
We specifically focus on the HalfCheetah-102D. Each entry of the weight matrix \(\mathbf{W}\) is constrained to the range \([-1, 1]\), and we conduct $5$ rollouts for each evaluation. To reformulate the original problem for constrained optimization, we ensure that the return from each rollout is greater than $-500$. This setup represents a high-dimensional modification of the Lunar Landing task as discussed in~\citet{eriksson2021scalable}.

 % which are satisfied if and only if the rover does not intersect any of these obstacles. In other words, any trajectory that avoids collision is feasible, and its constraint value is set to the minimum distance between the trajectory and the obstacle. Trajectories that pass through the obstacle incur a penalty, with more penalties applied the closer they come to the obstacle's center. We followed the implementation from \cite{wang2018batched}. \footnote{\url{https://github.com/zi-w/Ensemble-Bayesian-Optimization}} The code implementation of the constraints is not released from \cite{eriksson2021scalable}; we implement the constraint violation part. Below is a sample image of the best trajectory our method found.

 %figure part.

\subsection{Vehicle Design with 68 Constraints (MOPTA)}
MOPTA is the high-dimensional real-world problem of large-scale multidisciplinary mass optimization~\citep{anjos2009mopta}. The objective is to minimize a vehicle's mass, which incorporates decisions about materials, gauges, and vehicle shape with 68 performance constraints. The best-known optimum mass is approximately 222.74. We followed the implementation from \citet{papenmeier2022increasing}.\footnote{\url{https://github.com/LeoIV/BAxUS}}

\subsection{LassoBench}
LassoBench \citep{vsehic2022lassobench} \footnote{\url{https://github.com/ksehic/LassoBench}} is a high-dimensional benchmark for hyperparameter optimization, specifically designed to tune the hyperparameters of the Weighted LASSO (Least Absolute Shrinkage and Selection Operator) regression model. It includes both synthetic tasks (simple, medium, high, and hard) and real-world tasks (Breast cancer, Diabetes, Leukemia, DNA, and RCV1). In this work, we focus on the DNA task, a microbiology classification problem. It computes the average validation error across all cross-validation folds as an unconstrained objective. We reformulate the problem by retaining the mean validation error as the objective while introducing constraints that the validation error on each fold must not exceed 0.32.

% We also introduce the constraint that the validation error on each cross‑validation fold must not exceed 0.32.

\section{Baselines Details}
\label{app:baselines-details}
In this section, we provide a thorough description of our baseline implementation details and specify the hyperparameter settings used across all experiments.

\textbf{DiffOPT}
\citep{kong2025diffusion}:
% We reimplement this baseline using a network architecture comparable to our flow-based model, maintaining identical learning rates, activation functions, and optimizers for fair comparison.
As there is no publicly available code, we re-implement this baseline on our own. To approximate the data distribution, we use diffusion models with a similar architecture to our method. To enable accurate sampling from the target distribution, we implement Langevin dynamics as the energy function, which can be constructed by surrogate models in our setting, is differentiable.

\textbf{SCBO} \citep{eriksson2021scalable}: We follow the tutorial code for SCBO provided by \texttt{botorch} \footnote{\url{https://botorch.org/docs/tutorials/scalable_constrained_bo/}} to reproduce the results.

\textbf{PCAGP-SCBO} \citep{maathuis2024high}: To reproduce PCAGP-SCBO, we follow the code for SCBO and then apply \texttt{torch pca}\footnote{\url{https://github.com/valentingol/torch_pca}} to project high-dimensional data into a reduced latent space with dimension $l$ before fitting GP surrogates for constraints. For all synthetic tasks, we use $l=2$ and for real-world tasks, we conduct a hyperparameter search on $\left[2,  \lfloor D/2\rfloor\right]$ and report the best one.

\textbf{FuRBO} \citep{ascia2025feasibility}: We use the original implementation\footnote{\url{https://github.com/paoloascia/FuRBO}} for reproducing this baseline. FuRBO employs a feasibility-driven trust region strategy that adapts the search region based on constraint satisfaction.

\textbf{cEI} \citep{schonlau1998global}: We implement cEI acquisition function by using \texttt{qExpectedImprovement()} in \texttt{botorch} library. We train a GP surrogate model independently for the objective and each constraint.

\textbf{LogcEI} \citep{ament2023unexpected}: We implement logcEI acquisition function by using \texttt{qLogExpectedImprovement()} in \texttt{botorch} library. We train a GP surrogate model independently for the objective and each constraint.

\textbf{CMA-ES} \citep{hansen2006cma}: We employ the \texttt{pycma}\footnote{\url{https://github.com/CMA-ES/pycma}} library \citep{hansen2019pycma}. For constraint handling, we formulate the problem using the same Lagrangian approach with the same $\lambda$ value as ours for each task.

\textbf{COBYLA} \citep{powell1994direct}: We use the \texttt{scipy.optimize.minimize} implementation with \texttt{method='COBYLA'}. Constraints are passed directly to the optimizer in the form $-g^{(m)}(\mathbf{x}) \geq 0$ for each $m=1,\dots,M$.

\textbf{Sobol} \citep{sobol1967distribution}: We use Sobol quasi-random sequences generated via \texttt{SobolEngine()} as a non-adaptive space-filling baseline. At each step, the next point in the Sobol sequence is evaluated without any model-guided search.

\clearpage
\section{Implementation Details}
\label{app:implementation-details}
In this section, we introduce the implementation details of our method \textbf{CiBO}. Specifically, model architectures, the training processes employed, the hyperparameter configurations used, and the computational resources required.

\subsection{Training Models}
\subsubsection{Training Proxies}
We employ an ensemble of five proxies to model the objective function and a single proxy for each constraint. Each proxy is implemented as a MLP with three hidden layers of $1024$ units, using GELU \citep{hendrycks2016gaussian} activations. Proxies are trained with the Adam optimizer \citep{kingma2014adam} for $100$ epochs per round at a learning rate of $1 \times 10^{-3}$ and a batch size of $256$. All hyperparameters related to the proxy are listed in \Cref{table:proxy hyperparams}.
\begin{table*}[h]
\centering
\caption{Hyperparameters for Training Proxy}
% \resizebox{0.6\linewidth}{!}{
\begin{tabular}{c|ll}
\toprule
& Parameters & Values \\
\midrule
\multirow{3}{*}{Architecture} & Num Ensembles & $5$ \\
                              & Number of Layers & $3$ \\
                              & Num Units & $1024$ \\
\midrule
\multirow{4}{*}{Training} & Batch size & $256$ \\
                          & Optimizer & Adam \\
                          & Learning Rate & $1 \times 10^{-3}$ \\
                          & Training Epochs & $100$ \\
\bottomrule
\end{tabular}
% }
\label{table:proxy hyperparams}
\end{table*}

\subsubsection{Training Flow-based Models}
We adopt the architecture of \citet{lipman2024flow} for our flow model, comprising three hidden layers with $512$ units each. Training is performed using Adam optimizer for $500$ epochs per round, with a learning rate of \(1 \times 10^{-3}\) and a batch size of $256$. For ODE integration during sampling, we employ the Runge-Kutta $4$ method with \texttt{torchdiffeq} \citep{torchdiffeq}, and set the integration steps as $250$. All flow-model hyperparameters are detailed in \Cref{table:prior hyperparams}.
\begin{table*}[h]
\centering
\caption{Hyperparameters for Training Flow-based Model}
% \resizebox{0.6\linewidth}{!}{
\begin{tabular}{c|ll}
\toprule
& Parameters & Values \\
\midrule
\multirow{2}{*}{Architecture} & Number of Layers & $3$ \\
                              & Num Units & $512$ \\
\midrule
\multirow{4}{*}{Training} & Batch size & $256$ \\
                          & Optimizer & Adam \\
                          & Learning Rate & $1 \times 10^{-3}$ \\
                          & Training Epochs & $500$\\
\bottomrule
\end{tabular}
% }
\label{table:prior hyperparams}
\end{table*}
% as the backbone of our flow model. The architecture consists of three hidden layers, each containing 512 hidden units. During training, we use the Adam optimizer for 500 epochs per round with a learning rate of \(1 \times 10^{-3}\). We set the batch size to 256. We employ the Runge–Kutta method for numerical integration of the ODE in sampling. The hyperparameters related to the flow model are summarized in
% \cref{table:prior hyperparams}.
% \input{tables/flow}

% We utilize the temporal Residual MLP architecture from \citet{venkatraman2024amortizing} as the backbone of our diffusion model. The architecture consists of three hidden layers, each containing 512 hidden units. We implement GELU activations alongside layer normalization \cite{ba2016layer}. During training, we use the Adam optimizer for 50 epochs (100 for 400 dim tasks) per round with a learning rate of \(1 \times 10^{-3}\). We set the batch size to 256. We employ linear variance scheduling and noise prediction networks with 30 diffusion steps for all tasks. The hyperparameters related to the diffusion model are summarized in

% \input{tables/prior}

\clearpage
\subsection{Sampling Candidates}
\paragraph{Training Diffusion Sampler}
Various approaches have been developed to draw samples from a distribution when only an unnormalized probability density or energy function is available. Traditional methods include Markov Chain Monte Carlo (MCMC) techniques \citep{grenander1994representations, duane1987hybrid, Halton_1962, chopin2002sequential, skilling2006nested, lemos2024improving}, though their computational cost increases dramatically in high-dimensional spaces. More recently, amortized variational inference methods, those based on training diffusion samplers \citep{zhang2022path, vargas2023denoising, berner2024optimal, richter2024improved, vargas2024transport, lahlou2023theory, zhang2024diffusion}, have gained widespread adoption as they offer improved scalability for high-dimensional problems.

Following the~\citet{venkatraman2025outsourced}, we adopt method from~\citet{sendera2024improved} to train diffusion sampler to sample from the target:
\begin{equation}
    p_\text{post}(\mathbf{z}) \propto p(\mathbf{z})\exp\left(\beta\cdot r_\phi(f_{\theta}(\mathbf{z}))\right)
\end{equation}
Here, the right-hand-side term serves as an unnormalized probability density, which the diffusion sampler amortizes the sampling cost by approximating it.

\paragraph{Off-policy Training of Diffusion Sampler} As mentioned in the~\Cref{Method:Sampling Candidates}, we use the Trajectory Balance objective to train the diffusion sampler.
$$
\mathcal{L}_{\text{TB}}(\mathbf{z}_{0:1};\psi) = \left(\log  \frac{Z_\psi p(\mathbf{z}_0)\prod_{i=0}^{T-1} p_F(\mathbf{z}_{(i+1)\Delta t}|\mathbf{z}_{i \Delta t};\psi)}{p(\mathbf{z}_1)r(f_\theta(\mathbf{z}_1))\prod_{i=1}^Tp_B(\mathbf{z}_{(i-1)\Delta t}|\mathbf{z}_{i\Delta t})}\right)^2
$$
 The primary advantage of the TB loss is off-policy training~\citep{sendera2024improved, malkin2023gflownets}. We can train our model not only from the on-policy trajectories through the reverse SDE $\{\mathbf{z}_0, \cdots,  \mathbf{z}_1\}=\tau \sim p_F(\tau)$ but also from the trajectories through the forward SDE conditioned on the generated samples $\tau \sim p_B(\tau | \mathbf{z}_1)$. This proves its effectiveness on mode coverage and credit assignment~\citep{sendera2024improved}.

Specifically, we repeat two processes. First, we sample trajectories on-policy $\tau \sim p_F(\tau)$, train the model with \Cref{eq:posterior}, and collect the samples $\mathbf{z}_1$ into the buffer. Second, from the collected samples $\mathbf{z}_1$, we generate off-policy trajectories through $\tau \sim p_B(\tau|\mathbf{z}_1)$, then train with the \Cref{eq:posterior}. During the off-policy training, we prioritize the samples with low energy: $\mathcal{E}(\mathbf{z}_1) = -\log (p(\mathbf{z}_1)r(f_\theta(\mathbf{z}_1)))$ following~\citet{sendera2024improved} to make our model focus on the low energy samples. These techniques improve the overall performance of our framework (Appendix~\ref{app:policy}).

We use the original code\footnote{\url{https://github.com/GFNOrg/gfn-diffusion}} released from~\citet{sendera2024improved} for implementation. We also set method-specific hyperparameters with Path Integral Sampler (PIS)~\citep{zhang2022path} architecture, zero initialization, and t-scale to 1 to make sure the initialized $p_F(\mathbf{z}_1)$ starts from the standard normal distribution. Detailed hyperparameters for training the diffusion sampler can be found in \Cref{table:sampler hyperparams}.

\begin{table*}[h]
\centering
\caption{Hyperparameters for Training Diffusion Sampler}
% \resizebox{0.6\linewidth}{!}{
\begin{tabular}{c|ll}
\toprule
& Parameters & Values \\
\midrule
\multirow{3}{*}{Architecture} & Number of Layers & $2$ \\
                              & Num Units & $256$ \\
                              & Diffusion Time Steps & 50 \\
\midrule
\multirow{4}{*}{Training} & Batch size & $256$ \\
                          & Optimizer & Adam \\
                          & Learning Rate & $1 \times 10^{-3}$ \\
                          & Training Epochs & $50$ \\
\bottomrule
\end{tabular}
% }
\label{table:sampler hyperparams}
\end{table*}

\clearpage

\subsection{Hyperparameters}
%lambda, alpha, buffer, ... 기타등등
In our formulation of constrained black-box problems, we introduce $\lambda$ in \Cref{eq:Lagrangian Predictive Objective} for Lagrangian augmentation. We set $\gamma=1$, which is the exploration bonus term that controls the Exploration-Exploitation trade-off. Inverse temperature $\beta$ in \Cref{eq: unnormalized posterior} balances the influence between prior and likelihood, and is fixed to $10^5$. We draw $N\times B$ samples from the posterior distribution, then select $B$ samples during filtering, where we fix $N=10$ for all tasks. After evaluation, we update the training set by keeping the top $L$ highest‐scoring samples subject to the Lagrangian objective. We set $L=2000$ for synthetic tasks and $L=1000$ for real-world tasks. The reweighting hyperparameter is set to $\kappa=0.01$ for synthetic tasks and $\kappa=0.1$ for real-world tasks. The Lagrange multiplier $\lambda$ is crucial for balancing objectives and constraints, and we report the task-specific values in \Cref{table:candidate selection}. We include additional analysis to assess how each parameter affects overall performance in \Cref{sec:additional analysis} and Appendix~\ref{app:further-analysis}.
% We draw $10^2 \times B$ candidates for RoverPlanning and Mopta08,  $15^2 \times B$ candidates for DNA tasks from the diffusion sampler and select $B$ of them during filtering.
% For all tasks, we draw $10^2 \times B$ candidates from the diffusion sampler and select $B$ of them during filtering.

% For the upper confidence bound (UCB), we fixed $\gamma = 1.0$ that controls the exploration-exploitation.
% For the target posterior distribution, the inverse temperature parameter $\beta$ controls the trade-off between the influence of $\exp(r_\phi(\mathbf{x}))$ and $p_\theta(\mathbf{x})$.
% When selecting querying candidates, we sample $M=B \times 10^2$ candidates from $p_\psi(\mathbf{x})$, perform a local search for $J$ steps, and retain $B$ candidates for batched querying.
% % We maintain a fixed ratio of $M = 10^2 \times B$ across all tasks.
% After querying and adding candidates, we maintain our training dataset to contain $L$ high-scoring samples. We present the detailed hyperparameter settings in \cref{table:candidate selection}. We also conduct several ablation studies to explore the effect of each hyperparameter on the performance.
\begin{table}[ht]

\centering
\caption{Lagrange multiplier $\lambda$ for each task.}
\begin{tabular}{c c}\toprule
                       &Lambda $\lambda$
                       \\\midrule

Ackley 200D        &$10$\\
Rastrigin 200D        &$10$\\
Rosenbrock 200D        &$10$\\

\midrule

RoverPlanning 60D   &$5$\\
HalfCheetah 102D        &$5$\\
Mopta 124D        &$10$\\
DNA 180D            &$5$ \\\bottomrule

\end{tabular}
\label{table:candidate selection}
\end{table}

\clearpage

\section{Further Analysis}
\label{app:further-analysis}
In this section, we provide further analysis on different components of our method that are not included in the main manuscript due to the page limit.

\subsection{Analysis on Feasibility Ratio}
\label{app:feasibility}
To further analyze our method's ability to effectively handle constraints, we report the feasibility ratio across optimization batches for the Rastrigin 200D and RoverPlanning 60D tasks.

\begin{figure}[h]
    \centering
    \includegraphics[width=0.75\textwidth]{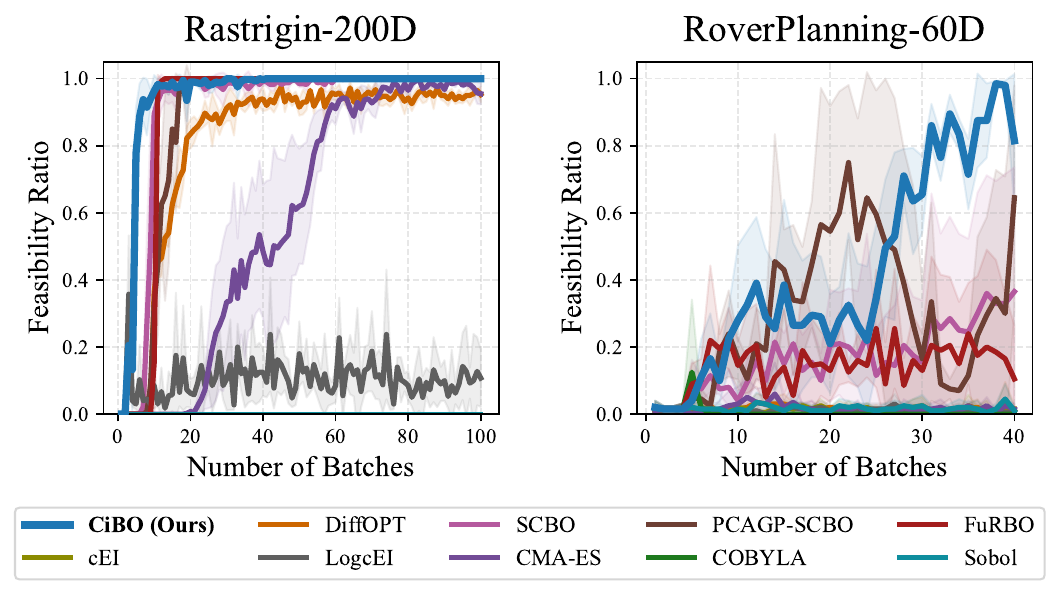}
    \caption {Feasibility ratio over all baselines. Experiments are conducted with four random seeds, and the mean and one standard deviation are reported.}
    \label{fig:feasibility}
    \vspace{-10pt}
\end{figure}

As illustrated in \Cref{fig:feasibility}, CiBO achieves the highest feasibility ratio within 5 to 10 batches, significantly faster than all competitors. While some baselines such as SCBO and PCAGP-SCBO eventually reach comparable ratios, they require about twice as many batches. In the Rover Planning tasks, most baselines fail to exceed a feasibility ratio of 0.4, whereas CiBO shows steady improvement and maintains high feasibility throughout the optimization process.

% As shown in \Cref{fig:feasibility}, CiBO demonstrates superior performance by rapidly achieving the highest feasibility ratio within the first 5-10 batches, significantly faster than all competing methods. While some baselines (SCBO, PCAGP-SCBO) eventually reach high feasibility ratios, they require approximately twice as many batches to achieve comparable performance. Other methods like DiBO and DiffOPT take even longer (around 20 batches), and CMA-ES struggles substantially, only reaching moderate feasibility ratios after 50 batches. Notably, CiBO not only reaches the high feasibility ratio faster but also maintains it consistently throughout the optimization process, demonstrating its robust constraint-handling capabilities in high-dimensional spaces.
% \clearpage

\subsection{Analysis on Reweighting Strategy}
\label{app:reweighting}
\begin{figure}[h]
    \centering
    \includegraphics[width=0.75\textwidth]{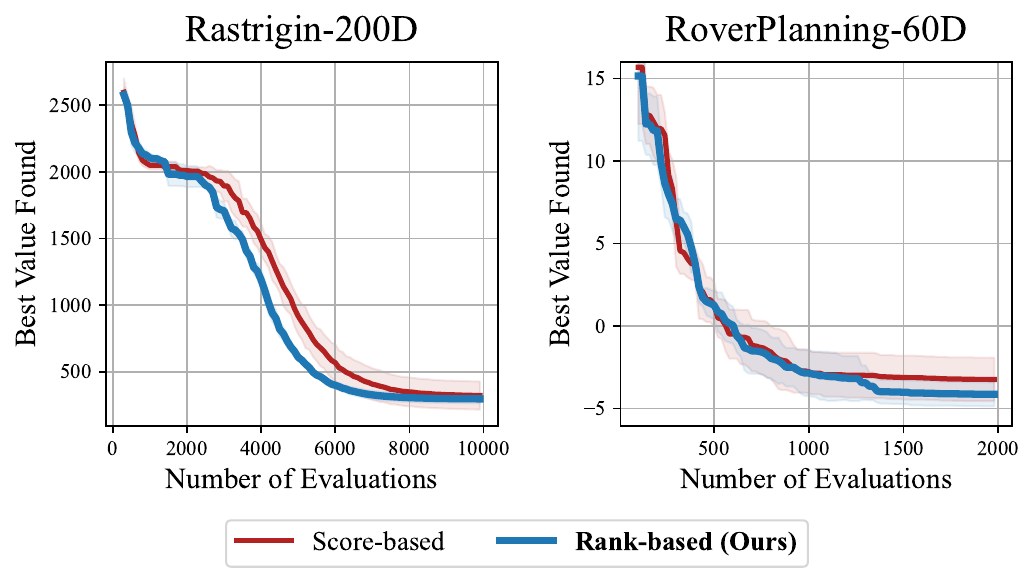}
    \caption{Comparison of reweighting strategies in Rastrigin-200D and Rover Planning-60D. Rank-based reweighting consistently outperforms score-based reweighting. Experiments are conducted with four random seeds, and the mean and one standard deviation are reported.}
    \label{fig:ablation_reweighting_strategy}
    \vspace{-20pt}
\end{figure}

We compare different reweighting strategies for training the generative model: score-weighted sampling that weights each sample proportionally to its Lagrangian-augmented value $w (y, c, D_r )$, and rank-based reweighting (ours) as described in \Cref{eq:weighted sampler}.
% Score-weighted sampling can be dominated by a few high-scoring outliers, especially in early rounds when the surrogate model is inaccurate. In contrast, rank-based reweighting provides a more stable training signal by using relative ordering rather than absolute values.
 As shown in \Cref{fig:ablation_reweighting_strategy}, rank-based reweighting outperforms score-based alternatives across both synthetic and real-world tasks.

\clearpage
\subsection{Effect of Off-policy Training in Amortized Inference}
\label{app:policy}
\begin{figure}[h]
    \centering
    \includegraphics[width=0.75\textwidth]{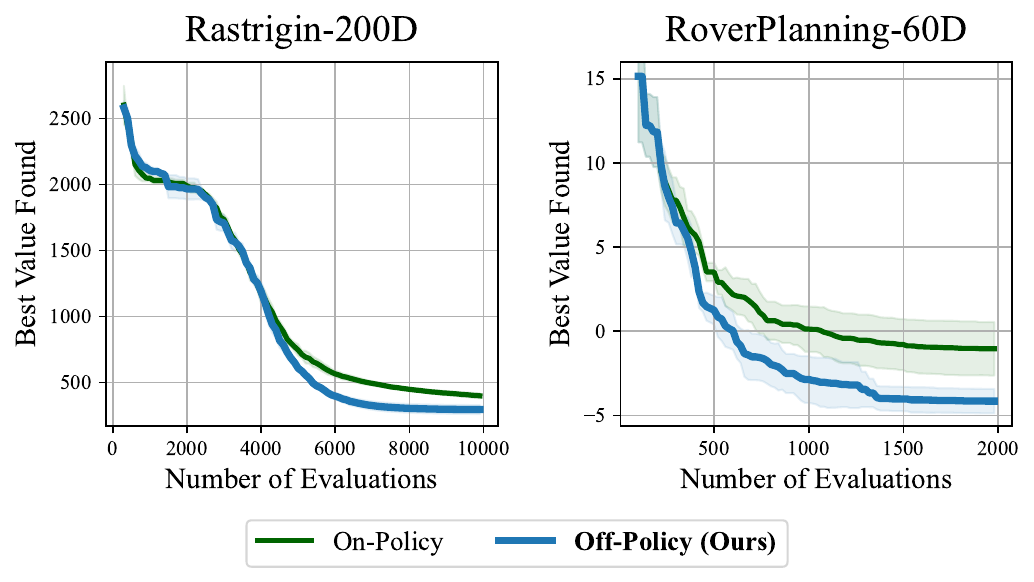}
    \caption {Comparison between off-policy and on-policy in  Rastrigin-200D and Rover Planning-60D. Experiments are
conducted with four random seeds, and the mean and one standard deviation are reported.}
    \label{fig:ablation_policy}
    \vspace{-10pt}
\end{figure}

We employ off-policy training with the TB loss to train the diffusion sampler.
To analyze the impact of off-policy training on performance, we conduct ablation studies on different training schemes.
As shown in \Cref{fig:ablation_policy}, off-policy training consistently outperforms on-policy methods, and the performance gap widens as the number of constraints grows (15 constraints in Rover Planning versus only 2 in Rastrigin). It highlights that training with off-policy samples is crucial for amortizing the posterior distribution, which has multiple modes and a large plateau, by improving mode coverage.

\subsection{Analysis on Adaptive $\lambda$}
\label{app:adaptive-lambda}
\begin{figure}[h]
    \centering
    \includegraphics[width=0.75\textwidth]{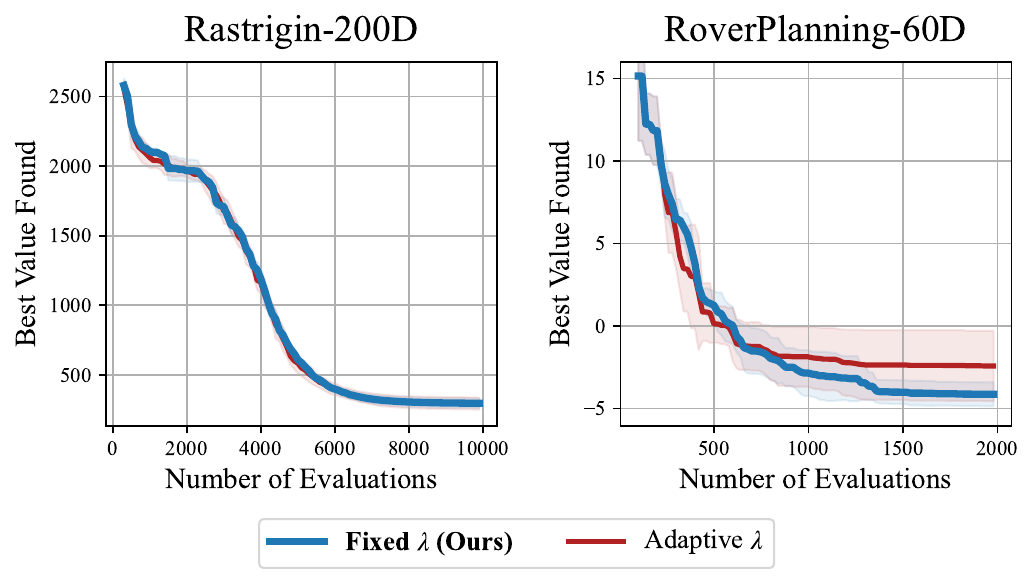}
    \caption{Comparison between fixed and adaptive $\lambda$ on Rastrigin-200D and RoverPlanning-60D. Experiments are conducted with four random seeds, and the mean and one standard deviation are reported.}
    \label{fig:ablation_adaptive_lambda}
    \vspace{-10pt}
\end{figure}

A natural question is whether adaptively updating the Lagrange multiplier $\lambda$ during optimization can improve performance, as is common in augmented Lagrangian methods. To investigate this, we compare our fixed $\lambda$ against an adaptive variant that updates $\lambda$ based on the observed constraint violations at each round. Specifically, we increase $\lambda$ when the feasibility ratio of the current batch is low and decrease it otherwise. As shown in \Cref{fig:ablation_adaptive_lambda}, the fixed $\lambda$ performs comparably to or better than the adaptive variant on both Rastrigin-200D and RoverPlanning-60D. On RoverPlanning, the fixed $\lambda$ achieves a notably better final value. We attribute this to the fact that adaptive updates introduce additional sensitivity to the noisy feasibility estimates in early rounds, whereas a fixed $\lambda$ provides a stable training signal throughout optimization. 
% Combined with the robustness to the specific choice of $\lambda > 0$ shown in \Cref{fig:abl-c}, these results justify our use of a fixed Lagrange multiplier.

\subsection{Analysis on Buffer Size $L$}
\label{app:buffer}
\begin{figure}[h]
    \centering
    \includegraphics[width=0.75\textwidth]{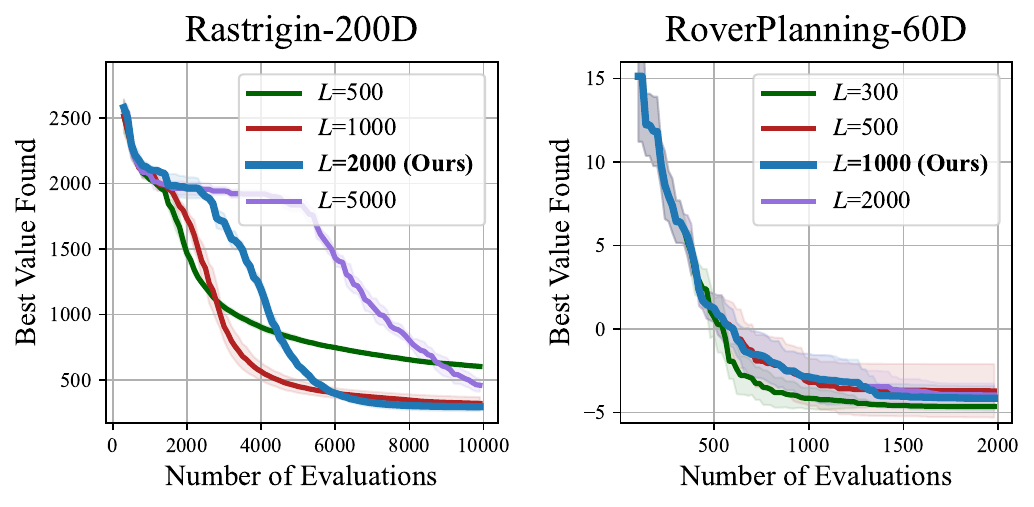}
    \caption {Performance of CiBO in Rastrigin-200D and Rover Planning-60D with varying $L$. Experiments are conducted with four random seeds, and the mean and one standard deviation are reported.}
    \label{fig:ablation_L}
    \vspace{-10pt}
\end{figure}

In each round, we retain the $L$ top-scoring samples with respect to the  Lagrangian-relaxed objective function for computational efficiency.
To analyze the effect of the buffer size $L$, we conduct experiments by varying $L$. As demonstrated in \Cref{fig:ablation_L}, using too small $L$ occasionally gets stuck in a sub-optimal solution, while using too large $L$ exhibits a slow convergence rate. Notably, the final performance of CiBO is not significantly influenced by the buffer size, indicating a low sensitivity to the choice of $L$.

% \clearpage

\subsection{Analysis on Initial Dataset size $|\mathcal{D}_0|$ and Batch size $B$}
\label{app:batch}
\begin{figure}[h]
    \centering
    \includegraphics[width=0.75\textwidth]{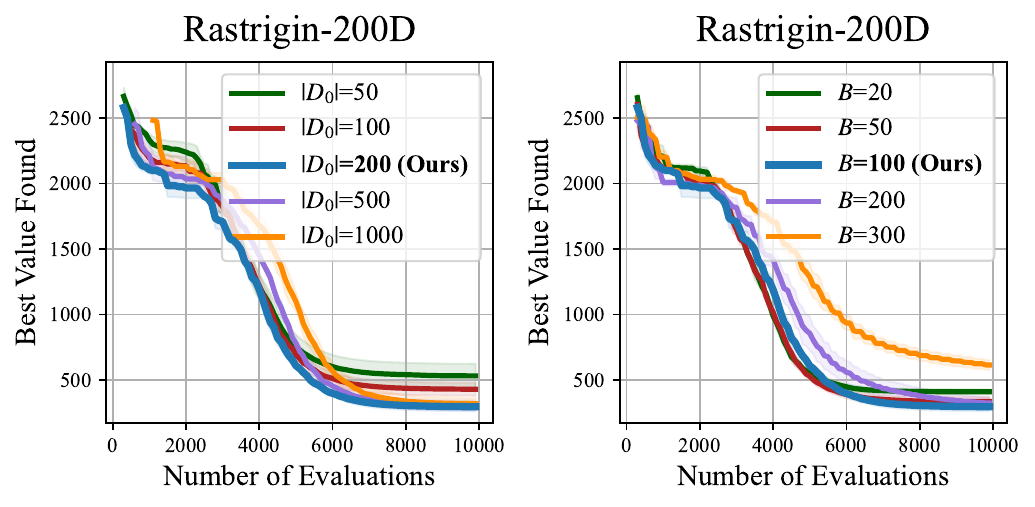}
    \caption {Performance of CiBO in Rastrigin-200D with varying $|D_0|$ and $B$. Experiments are conducted with four random seeds, and the mean and one standard deviation are reported.}
    \label{fig:ablation_DB}
    \vspace{-15pt}
\end{figure}

The size of the initial dataset, $|\mathcal{D}_0|$, and the batch size, $B$, play a critical role in the performance of black-box optimization algorithms.
When $|\mathcal{D}_0|$ is small and $B$ is large, the algorithm must optimize using very limited information, making the search significantly more challenging. To this end, we conduct experiments varying $|\mathcal{D}_0|$ and $B$ to demonstrate the robustness of our method on initial data configurations.
As shown in \Cref{fig:ablation_DB}, our method demonstrates robustness regarding both the initial dataset size $\vert\mathcal{D}_0\vert$ and the batch size $B$.

\clearpage
\subsection{Analysis on Misaligned Constraints}
\label{app:misaligned}

In the synthetic experiments, the optimal points of Rastrigin and Ackley are located at $\mathbf{x} = [0,0,...,0]$. Therefore, for those functions, directly optimizing the second constraint in Appendix~\ref{app:synthetic details} also optimizes the objective function. To verify that our method's performance does not simply benefit from this alignment between constraints and objectives, we conduct additional analysis on both the Rastrigin and Ackley functions with misaligned constraints: $\sum_{d=1}^{200}(x_d - 1) \leq 0$ and $||\mathbf{x}-[1,1,...,1]||_2^2\leq 30$, where the constraint centers are shifted away from the optimal point.

\begin{figure}[h]
    \centering
    \includegraphics[width=0.75\textwidth]{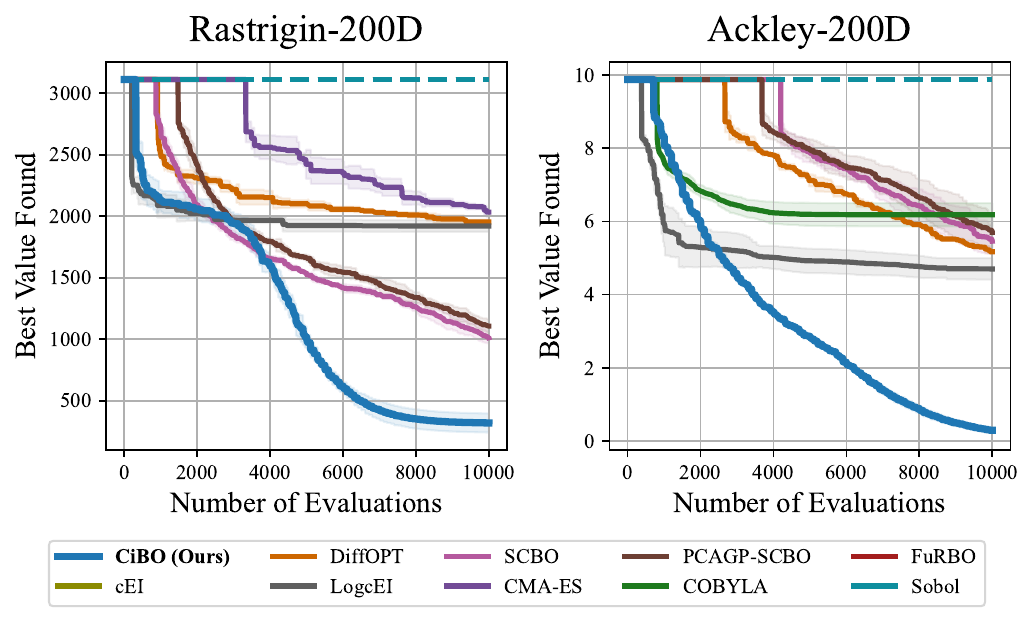}
    \vspace{-5pt}
    \caption {Analysis on misaligned constraints for Rastrigin-200D (left) and Ackley-200D (right). Experiments used four random seeds, reporting mean and one standard deviation.}
    \label{fig:misaligned}
    \vspace{-15pt}
\end{figure}

As shown in \Cref{fig:misaligned}, our method consistently outperforms the baselines on both tasks even with this misaligned constraint configuration. This demonstrates that the strong performance of our method stems from its expressive capacity to handle complex constraint landscapes, rather than benefiting from any particular alignment between constraint centers and objective optima.
% \clearpage

% \clearpage
\subsection{Comparison with Unconstrained Generative Methods}
\label{app:unconstrained}
\begin{figure}[h]
    \centering
    \includegraphics[width=0.75\textwidth]{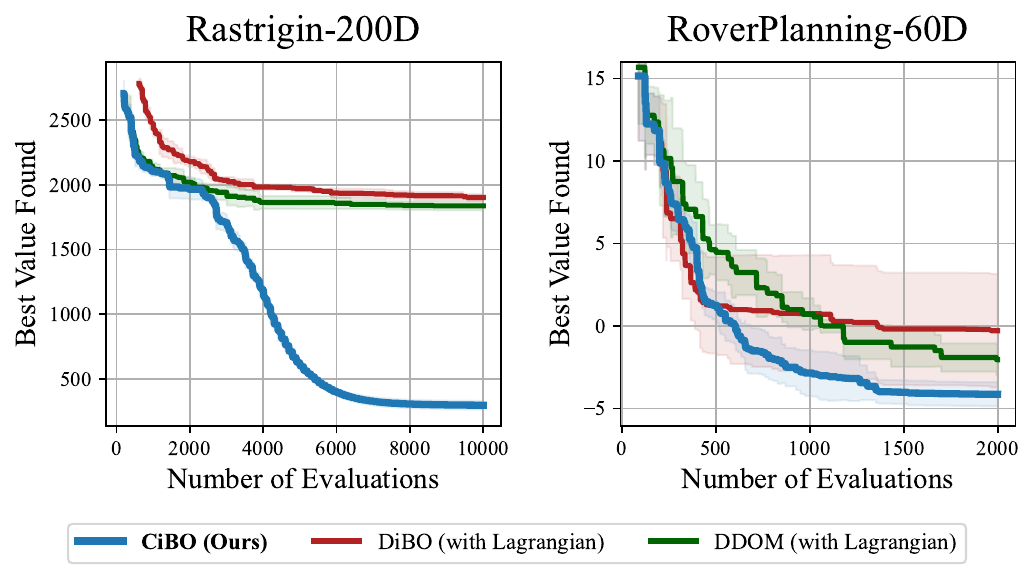}
    \caption{Comparison with unconstrained generative methods. Experiments are conducted with four random seeds, and the mean and one standard deviation are reported.}
    \label{fig:compare_dibo_ddom}
    \vspace{-20pt}
\end{figure}

DDOM~\citep{krishnamoorthy2023diffusion} and DiBO~\citep{yun2025posterior} are generative model-based methods originally designed for unconstrained black-box optimization. To enable a comparison in our constrained setting, we adapt both methods by incorporating Lagrangian relaxation, searching $\lambda \in \{1, 2, 5, 10, 20\}$ and reporting results with the best $\lambda$ per task. 
As shown in \Cref{fig:compare_dibo_ddom}, even with the best $\lambda$ selected per task, DiBO and DDOM both underperform CiBO. These results demonstrate that simply augmenting unconstrained methods with Lagrangian relaxation is insufficient for constrained settings, underscoring the need for constraint-aware optimization framework.

\section{Runtimes}
\label{app:Runtimes}
We report the per-round wall-clock time of each method in \Cref{tab:time_complexity_rastrigin,tab:time_complexity_rover}. We decompose each optimization round into four stages. \textit{Surrogate} measures the time spent fitting a surrogate model to the observed data (e.g., GP fitting for BO-based methods, proxy ensemble training for generative methods). \textit{Generative} captures the time for training a generative model that learns to propose candidates (applicable only to CiBO and DiffOPT). \textit{Candidates} is the time for proposing a batch of candidate solutions from the trained models or search procedure. \textit{Oracle} records the wall-clock time of evaluating the proposed candidates on the black-box objective and constraints. All experiments run on a single NVIDIA RTX 3090 GPU and Intel Xeon Platinum CPU @ 2.90\,GHz, pinned to 8 cores for fair comparison.

As shown in the tables, CiBO requires mostly less time per round than standard BO-based methods, because the lightweight proxy ensemble avoids the expensive GP fitting that dominates the runtime of those approaches.
\begin{table}[ht]
\centering
\caption{Per-round wall-clock time (seconds) on Rastrigin-200D. We report the mean $\pm$ one standard deviation over 4 seeds.}
\resizebox{\linewidth}{!}{
\begin{tabular}{l|cccc|c}\toprule
& Surrogate & Generative & Candidates & Oracle & Total \\
\midrule
cEI & $128.15 \pm 10.58$ & \text{---} & $11.82 \pm 0.16$ & $<$0.01 & $140.05 \pm 10.56$ \\
LogcEI & $127.82 \pm 10.06$ & \text{---} & $52.25 \pm 7.28$ & $<$0.01 & $180.12 \pm 6.08\phantom{0}$ \\
SCBO & $127.95 \pm 10.06$ & \text{---} & $3.70 \pm 0.00$ & $<$0.01 & $131.65 \pm 10.06$ \\
PCAGP-SCBO & $129.07 \pm 10.79$ & \text{---} & $3.70 \pm 0.00$ & $<$0.01 & $132.77 \pm 10.79$ \\
FuRBO & $127.85 \pm 10.09$ & \text{---} & $4.43 \pm 0.04$ & $<$0.01 & $132.35 \pm 10.09$ \\
\midrule
CMA-ES & \text{---} & \text{---} & $0.10 \pm 0.00$ & $<$0.01 & $\phantom{0}0.10 \pm 0.00$ \\
COBYLA & \text{---} & \text{---} & $0.10 \pm 0.00$ & $<$0.01 & $\phantom{0}0.10 \pm 0.00$ \\
Sobol & \text{---} & \text{---} & $0.10 \pm 0.00$ & $<$0.01 & $\phantom{0}0.10 \pm 0.00$ \\
\midrule
DiffOPT & $53.47 \pm 0.78$ & $12.05 \pm 0.05$ & $0.60 \pm 0.00$ & $<$0.01 & $66.12 \pm 0.83$ \\
\midrule
\textbf{CiBO} & $\mathbf{25.60 \pm 0.17}$ & $\mathbf{20.27 \pm 0.26}$ & $\mathbf{0.90 \pm 0.00}$ & $<$0.01 & $\mathbf{46.83 \pm 0.36}$ \\
\bottomrule
\end{tabular}}
\label{tab:time_complexity_rastrigin}
\end{table}

\begin{table}[ht]
\centering
\caption{Per-round wall-clock time (seconds) on RoverPlanning-60D. We report the mean $\pm$ one standard deviation over 4 seeds.}
\resizebox{\linewidth}{!}{
\begin{tabular}{l|cccc|c}\toprule
& Surrogate & Generative & Candidates & Oracle & Total \\
\midrule
cEI & $38.38 \pm 2.58$ & \text{---} & $25.32 \pm 0.31\phantom{0}$ & $0.60 \pm 0.00$ & $64.35 \pm 2.92\phantom{0}$ \\
LogcEI & $38.50 \pm 2.62$ & \text{---} & $34.25 \pm 14.12$ & $0.62 \pm 0.04$ & $73.38 \pm 15.01$ \\
SCBO & $38.75 \pm 2.72$ & \text{---} & $7.75 \pm 0.05$ & $0.60 \pm 0.00$ & $47.15 \pm 2.72\phantom{0}$ \\
PCAGP-SCBO & $10.25 \pm 1.82$ & \text{---} & $1.80 \pm 0.00$ & $0.60 \pm 0.00$ & $12.70 \pm 1.77\phantom{0}$ \\
FuRBO & $39.88 \pm 2.73$ & \text{---} & $8.00 \pm 0.00$ & $0.60 \pm 0.00$ & $48.55 \pm 2.77\phantom{0}$ \\
\midrule
CMA-ES & \text{---} & \text{---} & $<$0.01 & $0.60 \pm 0.00$ & $\phantom{0}0.60 \pm 0.00\phantom{0}$ \\
COBYLA & \text{---} & \text{---} & $0.10 \pm 0.00$ & $0.60 \pm 0.00$ & $\phantom{0}0.70 \pm 0.00\phantom{0}$ \\
Sobol & \text{---} & \text{---} & $0.10 \pm 0.00$ & $0.60 \pm 0.00$ & $\phantom{0}0.70 \pm 0.00\phantom{0}$ \\
\midrule
DiffOPT & $33.30 \pm 2.92$ & $2.40 \pm 0.00$ & $1.10 \pm 0.00$ & $0.62 \pm 0.04$ & $37.40 \pm 2.92\phantom{0}$ \\
\midrule
\textbf{CiBO} & $\mathbf{16.43 \pm 0.40}$ & $\mathbf{16.35 \pm 0.15\phantom{0}}$ & $\mathbf{0.90 \pm 0.00}$ & $0.60 \pm 0.00$ & $\mathbf{34.30 \pm 0.46}\phantom{0}$ \\
\bottomrule
\end{tabular}}
\label{tab:time_complexity_rover}
\end{table}

% \section{Broader Impact}
% \label{sec:Broader Impact}
% Advances in real-world design optimization have the potential to drive major innovations, but they also come with potential risks and unintended consequences. For example, optimization techniques in biochemical design may uncover novel compounds with therapeutic potential, but similar methods could also be misused to discover harmful substances. It is essential for researchers to act responsibly and ensure their work serves the public good.

%%%%%%%%%%%%%%%%%%%%%%%%%%%%%%%%%%%%%%%%%%%%%%%%%%%%%%%%%%%%%%%%%%%%%%%%%%%%%%%
%%%%%%%%%%%%%%%%%%%%%%%%%%%%%%%%%%%%%%%%%%%%%%%%%%%%%%%%%%%%%%%%%%%%%%%%%%%%%%%
% NEURIPS CHECKLIST
%%%%%%%%%%%%%%%%%%%%%%%%%%%%%%%%%%%%%%%%%%%%%%%%%%%%%%%%%%%%%%%%%%%%%%%%%%%%%%%
%%%%%%%%%%%%%%%%%%%%%%%%%%%%%%%%%%%%%%%%%%%%%%%%%%%%%%%%%%%%%%%%%%%%%%%%%%%%%%%
\newpage

\end{document}